\pdfoutput=1

\documentclass[11pt]{article}

\usepackage{EMNLP2023}

\usepackage{times}
\usepackage{latexsym}

\usepackage[T1]{fontenc}

\usepackage[utf8]{inputenc}
\usepackage{wrapfig,lipsum,booktabs}

\usepackage{microtype}
\usepackage{diagbox}
\usepackage{hyperref}       
\usepackage{url}            
\usepackage{booktabs}       
\usepackage{amsfonts}       
\usepackage{amsmath}
\usepackage{amssymb}
\usepackage{amsthm}
\usepackage{float}
\usepackage{graphicx}
\usepackage{paralist, tabularx}
\usepackage{bm}
\usepackage{nicefrac}       
\usepackage{xcolor}         
\usepackage{hhline}

\usepackage{booktabs}
\usepackage{multirow}
\usepackage{pifont}
\usepackage{setspace}
\usepackage{scrextend}
\usepackage{arydshln}
\usepackage{caption}
\usepackage{subcaption}
\usepackage{tikz}
\usepackage{tikz-dependency}

\usepackage{xspace}
\usepackage{makecell}

\newcommand{\floor}[1]{\lfloor #1 \rfloor}
\newcommand{\Mod}[1]{\ (\mathrm{mod}\ #1)}
\newcommand{\argmax}{\text{arg\,max}}

\newcommand{\newMethod}{\textbf{\texttt{Avg-K}}\xspace}
\newcommand{\generalTerm}{\text{S-FFN}\xspace}
\newcommand{\oracle}{\text{VanillaM}\xspace}
\newcommand{\smoe}{\text{SMoE}\xspace}
\newcommand{\oracleTerraformer}{\text{VanillaController}\xspace}
\newcommand{\lowRank}{\text{LoRKM}\xspace}
\newcommand{\pkm}{\text{PKM}\xspace}
\newcommand{\pkmFfn}{\text{PKM-FFN}\xspace}
\newcommand{\hash}{\text{RandHash}\xspace}
\newcommand{\switch}{\text{Switch}\xspace}
\newcommand{\baseline}{\text{Dense Baseline}\xspace}

\newcommand{\update}[1]{---\xspace}
\long\def\comment#1{}
\usepackage{inconsolata}
\usepackage{listings}

\definecolor{codegreen}{rgb}{0,0.6,0}
\definecolor{codegray}{rgb}{0.5,0.5,0.5}
\definecolor{codepurple}{rgb}{0.58,0,0.82}
\definecolor{backcolour}{rgb}{0.95,0.95,0.92}

\lstdefinestyle{mystyle}{
    backgroundcolor=\color{backcolour},   
    commentstyle=\color{codegreen},
    keywordstyle=\color{magenta},
    numberstyle=\tiny\color{codegray},
    stringstyle=\color{codepurple},
    basicstyle=\ttfamily\footnotesize,
    breakatwhitespace=false,         
    breaklines=true,                 
    captionpos=b,                    
    keepspaces=true,                 
    numbers=left,                    
    numbersep=5pt,                  
    showspaces=false,                
    showstringspaces=false,
    showtabs=false,                  
    tabsize=2
}

\title{Towards A Unified View of Sparse Feed-Forward Network in Pretraining Large Language Model}

\author{
Zeyu Leo Liu$^\diamondsuit$
\thanks{\;\;Work done as a Meta AI Resident, while studying at the University of Washington, Seattle. Correspondence to: Zeyu Leo Liu $<$\texttt{zliu@cs.utexas.edu}$>$} \\
\And
Tim Dettmers$^\spadesuit$ \\
\And
Xi Victoria Lin$^\heartsuit$  \\
$^\diamondsuit$The University of Texas at Austin  $^\spadesuit$University of Washington, Seattle $^\heartsuit$Meta AI\\
\And
Veselin Stoyanov$^\heartsuit$  \\
\And
Xian Li$^\heartsuit$ 
}

\begin{document}
\maketitle
\begin{abstract}
Large and sparse feed-forward layers (S-FFN) such as Mixture-of-Experts (MoE) have proven effective in scaling up Transformers model size for \textit{pretraining} large language models. By only activating part of the FFN parameters conditioning on input, S-FFN improves generalization performance while keeping training and inference costs (in FLOPs) fixed. 
In this work, we analyzed two major design choices of S-FFN: the memory block (a.k.a. expert) size and the memory block selection method under a general conceptual framework of sparse neural memory.
Using this unified framework, we compare several S-FFN architectures for language modeling and provide insights into their relative efficacy and efficiency. 
We found a simpler selection method --- \textbf{\texttt{Avg-K}} that selects blocks through their mean aggregated hidden states, achieving lower perplexity in language model pretraining compared to existing MoE architectures including Switch Transformer \cite{switch_transformer} and HashLayer \cite{hash_layer}.
\end{abstract}

\pdfoutput=1
\section{Introduction}
\label{sec:intro}
Large-scale pretrained language models (LLMs) achieve remarkable performance and generalization ability for NLP tasks \citep{gpt1, bert, roberta, gpt2, gpt3, t5, palm}. 
Scaling up the model size (the number of parameters)
has been shown as a reliable recipe for better generalization, unlocking new capabilities, while the performance has not shown signs of plateauing \citep{scaling_law_for_lm, opt,palm, palm_big, emergent_ability_of_large_language_models}.
However, the computational resources required to train larger language models are formidable, calling for more efficient training and inference solutions of LLMs \citep{retro,green_ai, tay2020efficient}.

One promising direction is sparse scaling which increases the number of parameters while keeping the training and inference cost (in FLOPs) fixed. Recent work focuses on scaling up a transformer's feed-forward network (FFN) with sparsely activated parameters, resulting in a scaled and sparse FFN (\generalTerm). 
There have been two major approaches to achieve \generalTerm. One treats \generalTerm as a neural memory \citep{end2end_memory_net} where a sparse memory retrieves and activates only parts of the memory cells \citep{pkm}. The other adopts Mixture-of-Expert Network (MoE) \citep{gshard, switch_transformer,glam, hash_layer, base_layer, x-moe} that replaces a single FFN module with multiple equal-sized ones (called ``experts") and only activates a few among many experts for a particular input.

\begin{figure*}[th]
\centering
 \includegraphics[width=1\linewidth]{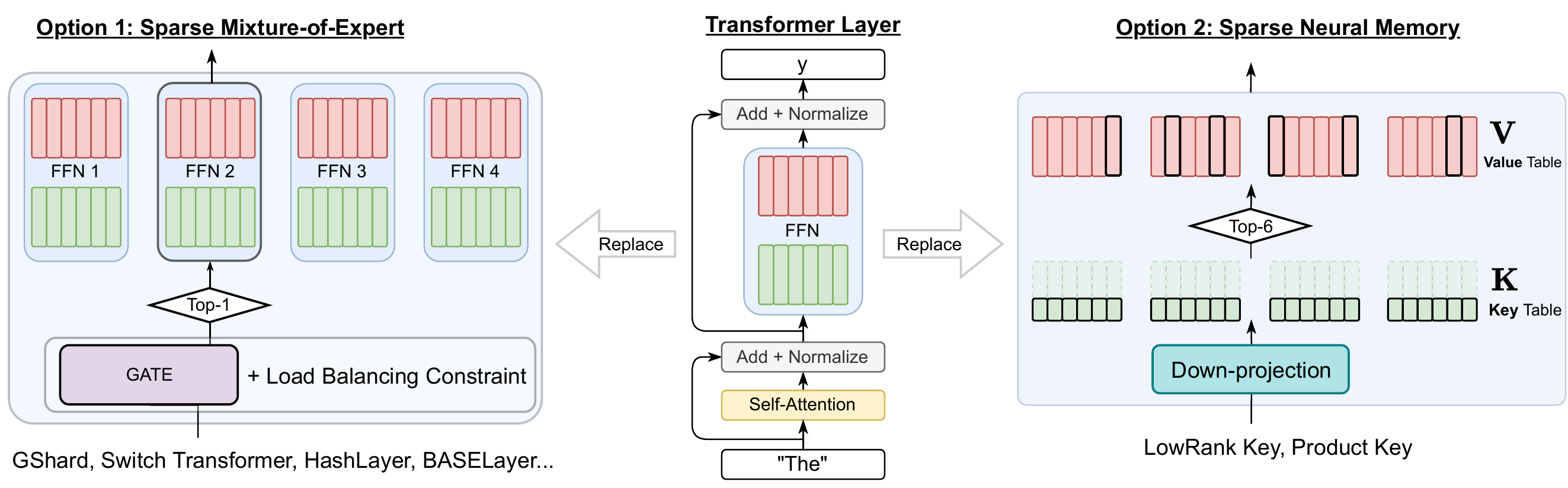}
\caption{Sparse Mixture-of-Expert and Sparse Neural Memory as two different methods.}
\label{fig:smoe+sparsememory}
\end{figure*}

While both memory and MoE models achieve \generalTerm, they have been considered two distinct approaches. We aim to draw the connections between these two classes of \generalTerm: What critical design choices do they have in common? Which design choices are essential for their modeling capability and computation efficiency? Can the effective ingredients of each method be transferred and combined to improve performance further? 

In order to answer these questions, we start from the neural memory view of FFN \citep{end2end_memory_net} (\textsection \ref{sec:bgd:ffn}) and reduce all \generalTerm's to the same mathematical form (\textsection \ref{sec:framework:moe_as_memory}). 
Then, we characterize these methods along two dimensions --- memory block size (e.g. expert size) and memory block selection method (e.g. gating) (\textsection \ref{sec:framework:main}).
Using this framework, we made the following contributions:
\begin{itemize}
    \item  We study a wider range of memory block sizes besides the commonly used block size in MoE architectures \citep{moe_survey_2022} and show that 
    reducing the block size keeps improving the perplexity with little incurred extra FLOPs (\textsection\ref{sec:framework_analysis:granularity} \textsection\ref{sec:framework_analysis:selection}),  leading to better perplexity/computation trade-offs.
    \item We conduct a systematic exploration of block selection methods to quantify their relative efficacy and efficiency (\textsection\ref{sec:framework_analysis:selection}). Specifically, we find that the selection method through a gating function, in general, improves the FLOPs-Perplexity trade-off. However, the parameterization of the current MoE's \cite{switch_transformer, base_layer} gating function --- multiplying token representation with a separately learned matrix --- has worse perplexity than using the FFN hidden states --- multiplying token representation with FFN's matrix (\textsection \ref{sec:bgd:ffn}).

    \item Drawing on the insights above, we propose a simple gate for \generalTerm --- \newMethod (\textsection \ref{sec:new_routing}) as a hybrid design choice between sparse neural memory and a mixture of experts. It efficiently selects memory blocks based on the mean aggregated hidden states of each block. 
    With 1\% additional FLOPs, \newMethod achieves lower perplexity (14.80) on the validation set of The Pile \cite{pile} than both Switch Transformer (16.45) \cite{switch_transformer} and HashLayer (15.75) \cite{hash_layer}. Moreover, \newMethod is the first MoE model that performs well without load balancing constraint, while conventional MoE transformers like Switch Transformer will degenerate \citep{switch_transformer, first_modern_moe, factored_representation_moe}.
\end{itemize}
\pdfoutput=1
\section{Background}
\label{sec:bgd}
\subsection{Feed-Forward Network}
\label{sec:bgd:ffn}
A transformer layer \citep{vanilla_transformer} consists of a self-attention block and a Feed-Forward Network block (FFN). FFN receives an input vector $\mathbf{x} \in \mathbb{R}^d$ from the self-attention block, multiplies it with parameter matrix $\mathbf{K} \in \mathbb{R}^{d_m \times d}$, applies a non-linear function $f$ to obtain the hidden states $\mathbf{m} \in \mathbb{R}^{d_m}$ and applies another affine transformation $\mathbf{V} \in \mathbb{R}^{d_m \times d}$ to produce a $d$-dimensional output $\mathbf{y}$. 
This multi-layer perceptron architecture can be expressed as:
\begin{align}
    \mathbf{y} = \text{FFN}(\mathbf{x}) = f(\mathbf{x} \cdot \mathbf{K}^\top) \cdot \mathbf{V} = \mathbf{m} \cdot \mathbf{V}. \label{eq:ffn_mlp}
\end{align}
Additionally, we could also view it as a neural memory \citep{sukhbaatar2015end, sukhbaatar2019augmenting, key_value_memory} (Eq. \ref{eq:ffn_memory}).
\begin{align}
    \mathbf{y} = \sum_{i=0}^{d_m-1} f(\mathbf{x} \cdot \mathbf{k}_i) \cdot \mathbf{v}_i = \sum_{i=0}^{d_m-1} m_i \cdot  \mathbf{v}_i. \label{eq:ffn_memory}
\end{align}
In this view, FFN consists of $d_m$ key-value pairs, known as \textit{memory cells}. Each key is represented by a $d$-dimensional $\mathbf{k}_i \in \mathbb{R}^d$, and together form the \textit{key table}; likewise, value vectors $\mathbf{v}_i \in \mathbb{R}^d$ constitutes a \textit{value table}  $\mathbf{V} \in \mathbb{R}^{d_m \times d}$. The memory multiplies the query input $\mathbf{x} \in \mathbb{R}^d$ with every $\mathbf{k}_i$; followed by the non-linear function, it produces \textit{memory coefficient} $m_i = f(\mathbf{x} \cdot \mathbf{k}_i)$ for the $i$-th memory cell. Finally, the output of FFN is the sum of its values $\mathbf{v}_i$ weighted by their corresponding memory coefficient $m_i$. Conventionally, the \textit{size} of FFN --- $d_m$ --- is set to be $4 \cdot d$.

\subsection{Scaling up FFN}
\label{sec:bgd:larger_ffn}
As discussed in \textsection \ref{sec:intro}, scaling up the number of parameters in FFN serves as a lever to improve transformer performance. Since a standard FFN accounts for about two-thirds of a transformer layer's parameters \citep{key_value_memory}, scaling up FFN will greatly affect the parameter size of a transformer model. However, one could sparsely activate the parameter to control the required compute. In this section, we 
and discuss two approaches to achieve a \textit{scaled} and \textit{sparse} FFN (\generalTerm). One has a mixture-of-expert model activate a few experts (\textsection \ref{sec:bgd:moe}), and the other specifies a memory model to sparsify (\textsection \ref{sec:bgd:non-moe}). 


\subsubsection{Mixture of Experts (MoE)}
\label{sec:bgd:moe}
Mixture of experts (MoE;  \citet{original_moe}) consists of a set of expert models $\left\{ \mathbf{f}_i(\mathbf{x})\right\}_{i=0}^{B-1}$ and a gating function $\mathbf{g}: \mathbb{R}^d \rightarrow \mathbb{R}^B$ to estimates the relevance of each expert. Finally, the output is the sum of experts' output weighted by the gate's weight estimation for that particular expert. 
\begin{align}
    \text{MoE}(\mathbf{x}) = \sum_{i\in \mathcal{E}= \left\{0, 1, \cdots, B-1\right\}} \mathbf{g}_i(\mathbf{x}) \cdot \mathbf{f}_i(\mathbf{x}) \label{eq:moe}
\end{align}
Recent work \citep{glam, gshard, hash_layer, base_layer, moe_by_topk_token} have applied this approach to transformer by dissecting the FFN into 
multiple expert blocks and \textbf{s}parse activation of the \textbf{MoE} (\textbf{\smoe}).
In \smoe, the gating function (or ``router'') routes an input token $\mathbf{x}$ to a subset\footnote{ Conventionally, \smoe implements \textbf{load balancing} constraints to prevent the overuse of certain experts and the under-utilization of others, and avoid convergence to local optima \citep{first_modern_moe, factored_representation_moe}.} (e.g. 1 or 2) of experts, $\mathcal{E} = \text{subset}(\mathbf{g}(\mathbf{x}))$. 
Previous work mainly adopts two types of gates.
\paragraph{Learned gate} \hspace{-1em} is parameterized by a set of learnable expert embeddings $\bm{\theta} = \left[\mathbf{e}_0; \cdots ; \mathbf{e}_{B-1}\right] \in \mathbb{R}^{B \times d}$, 
where each embedding corresponds to one expert.
The 
relevance of the $i$-th expert is obtained by 
$$\mathbf{g}_i(\mathbf{x}) = \frac{\exp\left(\mathbf{e}_i \cdot \mathbf{x}\right)}{\sum_j \exp\left(\mathbf{e}_j \cdot \mathbf{x}\right)}$$

To enforce load balancing when routing, 
previous work have employed an additional auxiliary loss \citep{gshard, switch_transformer, fb_moe, glam, x-moe} or framed expert utilization as a constrained optimization problem \citep{base_layer, moe_by_topk_token}.
\paragraph{Static gate}\hspace{-1em}, in contrast to a learnable gate, does not have any differentiable parameters. Instead, it uses a static mapping that encodes load-balancing constraints to route input \citep{hash_layer, demix}.
For example, 
\hash from HashLayer \citep{hash_layer} uses a hash table that maps from token type to \textit{randomly} selected expert(s). 
DEMix \citep{demix} ensures 
each expert only sees data from 
a pre-defined domain.

\subsubsection{Sparse Neural Memory}
\label{sec:bgd:non-moe}

The other line of work follows the memory view of FFN (Eq. \ref{eq:ffn_memory}). It is straightforward to increase the memory size $d_m$ to a much larger value $d_m \gg 4 \cdot d$. 
By only using the top-$k$ entries in the memory coefficient $\mathbf{m} = \mathbf{x} \cdot \mathbf{K}^\top$ (Eq. \ref{eq:ffn_memory}), one could sparsely activate the value table, resulting in a vanilla sparse memory (\textbf{\oracle}). 
However, the 
naive implementation of this approach results in 
computation cost proportional \textit{linearly} to the memory size. 
\citet{pkm} explored the following two techniques in this direction to scale computation sublinearly.

\paragraph{Low-Rank Key Memory (\lowRank)} 

A straightforward technique is to assume that the full key table $\mathbf{K}^\top \in \mathbb{R}^{d\times d_m}$ is composed of and approximated by a downward projection $\mathbf{D} \in \mathbb{R}^{d \times d_\ell}$ and a low-rank key table $\tilde{\mathbf{K}} \in \mathbb{R}^{d_m \times d_\ell}$, 
\begin{equation*}
    \mathbf{K}^\top = \mathbf{D}  \cdot  \tilde{\mathbf{K}}^\top
\end{equation*}
where $d_\ell \ll d$.
\lowRank produces memory coefficients by $m_i = f((\mathbf{x} \cdot \mathbf{D}) \cdot \tilde{\mathbf{k}}_i) = f(\mathbf{t} \cdot \tilde{\mathbf{k}}_i)$.
\paragraph{Product Key Memory (\pkm)} Building upon \lowRank, \pkm further decomposes the low-rank key table by assuming different low-rank keys have structured sharing with each other. See Appendix \ref{appendix:pkm} for more technical details. Due to such factorization, \pkm has a negligible key table $\mathbf{K}^\top = \mathbf{D} \cdot \tilde{\mathbf{K}}^\top$ (e.g., $<0.3\%$) relative to the parameters in the value table.
\pdfoutput=1
\section{A Unified View of Sparse FFNs}
\label{sec:framework}

We show the connections between MoE and neural memory despite their different formulations on the surface. We first derive 
a variant form of MoE to establish its connection with sparse memory (\textsection \ref{sec:framework:moe_as_memory}). Then, we propose a unified framework for \generalTerm (\textsection \ref{sec:framework:main}).
\subsection{A Neural Memory View of MoE}
\label{sec:framework:moe_as_memory}
MoEs use a gating function to estimate the importance of all experts and combine each expert's output through linear combination. Here, inspired by the memory view on FFNs (\textsection \ref{eq:ffn_memory}), we could view MoE as a wide neural memory chunked into $B$ FFNs:
\begin{align}
    & \text{MoE}(\mathbf{x}) 
    = \sum_{i=0}^{B-1} \mathbf{g}_i(\mathbf{x}) \cdot \text{FFN}^{(i)}(\mathbf{x}) \nonumber \\
    & = \sum_{i=0}^{B-1} \mathbf{g}_i(\mathbf{x}) \cdot \left(\sum_{j=0}^{d_m-1} m_j^{(i)} \cdot \mathbf{v}_j^{(i)}\right) \nonumber \\
    &= \sum_{i=0}^{B-1} \sum_{j=0}^{d_m-1} \left( \mathbf{g}_i(\mathbf{x}) \cdot m_j^{(i)}\right) \cdot \mathbf{v}_j^{(i)} \nonumber  \\
    &= \sum_{l=0 \text{ s.t. } l = i \cdot d_m + j}^{B \cdot d_m-1} \left(\mathbf{g}_{i}(\mathbf{x}) \cdot m_{j}^{(i)} \right) \cdot \mathbf{v}_{j}^{(i)} \nonumber \\
    & = \sum_{l=0}^{B \cdot d_m-1} m_l \cdot \mathbf{v}_{l},  \label{eq:moe-mem:final} 
\end{align}
where $\text{FFN}^{(i)}$ denotes the $i$-th FFN.

Based on linear algebra, we are able to write the standard MoE formulation (\textsection \ref{sec:bgd:moe}) in a similar summation form as that of neural memory (Eq \ref{eq:moe-mem:final}; \ref{sec:bgd:ffn}). MoE in its neural memory form has a
value table $\mathbf{V} = \left[\mathbf{V}^{(0)}; \cdots ; \mathbf{V}^{(B-1)}\right]\in \mathbb{R}^{B \cdot d_m  \times d}$ --- the concatenation of $B$ value tables $\mathbf{V}^{(i)} \in \mathbb{R}^{d_m  \times d}$ from all $\text{FFN}^{(i)}$. In this view, $\mathbf{v}_{l}$ with $l = i \cdot d_m + j $ corresponds to the $j$-th value vector in the $i$-th chunk of value table $\mathbf{V}^{(i)}$. Thus, its corresponding memory coefficient $m_{l} = \mathbf{g}_i(\mathbf{x}) \cdot m_j^{(i)}$ is produced by weighting the $j$-th memory coefficient of $\text{FFN}^{(i)}$, $m_j^{(i)}=\mathbf{x}\cdot\mathbf{k}_j^{(i)}$, by the relevance score of $\text{FFN}^{(i)}$, $\mathbf{g}_i(\mathbf{x})$. 
Through this memory view, one could see that a \textit{sparse} $\text{MoE}(\mathbf{x})$ is a \textit{sparse} memory operating in terms of $B$ memory blocks; it uses the gate $\mathbf{g}(\mathbf{x})$ to narrow down the calculation over the stacked value tables $\mathbf{V}$ to the value tables from $\text{FFN}^{(i)}, \text{for }i \in \text{subset}(\mathbf{g}(\mathbf{x}))$ (i.e. sparsify).
\paragraph{Comparison with Sparse Memory} Both \smoe and sparse neural memory are neural memory, but there are several differences: 
\textbf{1)} whether memory cells share the same relevance weight: in sparse neural memory, each memory cell receives an individual weight $m_i$. In contrast, in \smoe, each group of $4 \cdot d$ memory cells shares the same relevance weight $\mathbf{g}_i(\mathbf{x})$.
\textbf{2)} memory selection criterion: if we center the key-value computation (Eq. \ref{eq:ffn_memory}) at the core of \generalTerm computation, the sparse memory directly uses the memory parameters for selection --- the dot product between input token $\mathbf{x}$ and key vectors $\mathbf{k}_i$, whereas \smoe depends on a separately parameterized gate $\mathbf{g}$. 



\subsection{The Unified Framework}
\label{sec:framework:main}
We propose a general framework that unifies the two different approaches to achieve \generalTerm. We view both as instances of a memory with large key and value table --- $\mathbf{K} \in \mathbb{R}^{d_m \times d_k}, \mathbf{V} \in \mathbb{R}^{d_m \times d_v}$, where $d_m \gg 4 \cdot d$. We distinguish the different methods along two dimensions illustrated below and summarized in Table \ref{tab:methods-summary}:
\pdfoutput=1
\begin{table*}[t]
\renewcommand{\arraystretch}{1.25}
\setlength{\tabcolsep}{4pt}
\small
\centering

\begin{tabular}{c|ccccc|c}
\toprule
\begin{tabular}[c]{@{}c@{}}
    \textbf{Memory block} \\
    \textbf{size} ($g$)
\end{tabular}  & \multicolumn{4}{c}{\textbf{Memory block selection method}}  & &\textbf{Model Name} \\
\hline
\multirow{2}{*}{1} & & \multirow{2}{*}{Direct} & & Full-parameter Key  & & \oracle  \\
\cline{5-7}
                   & &  & & Low-rank Key & & \lowRank, \pkm  \citep{pkm}\\
\hline
 \multirow{3}{*}{$4 \cdot d$}            & & \multirow{3}{*}{Indirect} & & Learned gate & &
 \begin{tabular}[c]{@{}c@{}}
    Switch Transformer\citep{switch_transformer}, \\
    GShard \citep{gshard},\\
    GLaM \citep{glam}, \\ 
    BASELayer \citep{base_layer}, \\
    X-MoE \citep{x-moe} \\
\end{tabular}    \\
\cline{5-7}
  & &  &  & Static gate   & &
 \begin{tabular}[c]{@{}c@{}}
    HashLayer\citep{hash_layer},\\
    DEMix \citep{demix} 
\end{tabular}  \\ 
\bottomrule
\end{tabular}
\caption{\generalTerm methods decomposed along the defined design dimensions.}
\label{tab:methods-summary}
\end{table*}

\paragraph{Memory block size} \hspace{-1em} specifies how many memory cells share the same relevance weight at selection time, and thus together treated as a memory \textit{block}. We use $g$ to denote the size of one block. In other words, we split the $\mathbf{K}, \mathbf{V}$ along the $d_m$-dimension into $g$-size blocks. Therefore, a memory consists of $B = d_m/g$ blocks in total. Formally, we write

\begin{align*}
    \mathbf{K}^{g} & = \left[\mathbf{K}^{(0)};  \mathbf{K}^{(1)} ; \cdots ; \mathbf{K}^{(B-1)}\right] \in \mathbb{R}^{d_m \times d_k}\\
    \mathbf{V}^{g} & = \left[\mathbf{V}^{(0)};  \mathbf{V}^{(1)} ; \cdots ; \mathbf{V}^{(B-1)}\right]  \in \mathbb{R}^{d_m \times d_v} 
\end{align*}

For example, sparse memory has block size $g=1$ --- trivially treating 1 memory cell as a ``block''; and \smoe has the block size $g=4 \cdot d$ (\textsection \ref{sec:framework:moe_as_memory}). Current approaches generally use fixed block sizes, but this is mostly an artifact of how the methods were derived rather than a mathematical constraint. For example, we can design \smoe versions instead of $1$ expert of size $4 \cdot d$, or uses $2$ experts of size $2 \cdot d$. We can similarly chunk memory coefficients $\mathbf{m}$ into blocks of size $g$ 
in sparse memories.

\paragraph{Memory block selection method} \hspace{-1em} is the specific function that compute the relevance of each memory blocks for selection. Since \smoe is also a type of sparse memory, we distinguish the selection method by a new criterion --- \textit{whether one allows input $\mathbf{x}$ to \textit{directly interact} with the key table $\mathbf{K}^g$}. As discussed in \textsection \ref{sec:framework:moe_as_memory}, \smoe uses the estimation from an individually parameterized gate to select, while sparse memory solely and directly uses a key table. Thus, current \smoe is a type of \textit{indirect} selection method, and sparse memory a \textit{direct} one. Various SMoEs are further characterized by whether their gating function has learned parameters or consists of a static mapping (\textsection\ref{sec:bgd:moe}). Meanwhile, sparse memory is characterized by how much factorization the key table uses (\textsection\ref{sec:bgd:non-moe}). 


\subsection{A New Selection Method --- \newMethod}
\label{sec:new_routing}
The gate design in the Mixture-of-Expert methods ensures that not all experts are activated for routing tokens. Without the gate design, conventional sparse neural memory methods (\S\ref{sec:bgd:non-moe}) uses the full key table before sparsifying the computation in the value tables, which explains the increasing computation cost for scaling up sparse neural memory. As later shown in Fig. \ref{fig:355m_pareto} (\S \ref{sec:framework_analysis:granularity}), this additional computation doesn't bring proportional improvement to its performance, compared with MoE. 

However, there is still merit in sparse neural memory that MoE could acquire. Based on the contrastive analysis from \textsection\ref{sec:framework_analysis:selection}, we found that when the model \textit{makes more use of each expert's key table} for routing tokens, the more its performance will increase. This is precisely the deficiency in the conventional Mixture of Experts (MoE) model, as it relies on a separately learned parameter or a static gating mechanism (\textsection \ref{sec:bgd:moe}). 

To this end, we propose a new routing method --- \newMethod  --- as a hybrid design choice between sparse neural memory and MoE methods. 
\newMethod represents each block with the average of its key table $\mathbf{K}^{(i)} \in \mathbb{R}^{g \times d}$ along $g$-dimension:
$$\mathbf{e}_i = \frac{1}{g} \cdot \sum_{j=0}^{g-1} \mathbf{k}^{(i)}_j = \texttt{\textbf{Avg}}(\mathbf{K}^{(i)}\texttt{, dim=0})$$

Then, we use the dot product between $\mathbf{x}$ and the averages to select the top-$b$ selected block and route the token there for memory calculation (Eq. \ref{eq:ffn_memory}):
$$\mathbf{g}_i(\mathbf{x}) =
    \begin{cases}
      1 & i \in \left\{\text{top-}b \text{ of }[\mathbf{e}_0 , \cdots, \mathbf{e}_{B-1}] \cdot \mathbf{x}\right\} \\
      0 & \text{otherwise}
    \end{cases}\label{eq:new_method:route}$$

Due to the linearity of averaging, the operation $\mathbf{e}_i \cdot \mathbf{x}$ is equivalent to calculating the average of dot products within a block without GeLU. Since all tokens share the set of averaged key vectors, our method is efficient. In summary, \newMethod migrates MoE's advantages: a gate design, and a full-parameterized key table in each expert. As an advantage of sparse memory, it uses the average key vectors of each expert to route tokens and thus increases the memory selection method's dependency on the key table. We provide more rationale for our choice of average function in Appendix \ref{appendix:newMethod:rationale}.

\label{sec:sparsity-clss}

\pdfoutput=1
\section{Experiment Setup}
\label{sec:exp}

\subsection{Models}
\label{sec:exp:model}
We choose \textbf{\baseline} using transformer architectures used in GPT-3 models \citep{gpt3}, which has 24 transformer layers, with $d=1024$, $f = \text{GeLU}$ as activation function, and with a memory size (or FFN hidden size) to be $4 \cdot d$. This model is also referred to as the "base model" because S-FFN's size is chosen based on the configuration of this model. Our choice of architecture size leads to a base model with 355M parameters (See Appendix \ref{appendix:setting}). Our reason for the chosen model size is two-fold: 1) this size is similar to the community-estimated size of OpenAI \texttt{text-ada-001}; 2) as indicated in \citet{scaling_law_for_smoe}, 355M is the smallest size that separates the performances of different architecture designs.

\paragraph{\generalTerm} Given a model above, we replace some of its FFNs with an \generalTerm. Similar to \citep{gshard}, we replace the FFN at every 6 layers (layer $5, 11, 17, 23$, indexed from 0), leading to 4 S-FFNs in total across 24 layers. We use $k$ to denote the number of memory blocks used and control how activated the \generalTerm is. We use the formulation of $d_m = E \cdot (4\cdot d)$ to control the size of \generalTerm, so the \generalTerm will activate $b = \frac{k}{g}$ out of $B = \frac{d_m}{g}$ memory blocks. In Table \ref{tab:analysis_experiment}, we list all \generalTerm models used for analysis in \textsection \ref{sec:framework_analysis}. We count FLOPs analytically following \citet{large_scale_with_megatron_lm} and do not account if a worker finishes computation before another (when using model parallelism). We use \textit{the number of learnable parameters} to consider whether two models are equally expressive. 
In Table \ref{tab:analysis_experiment}, we list all \generalTerm models used for analysis in \textsection \ref{sec:framework_analysis}. For experimentation of our \newMethod, we don't enforce load balancing for simplicity. 

\pdfoutput=1
\begin{table*}[ht]
\renewcommand{\arraystretch}{1.25}
\setlength{\tabcolsep}{2pt}
\small
\begin{center}
\begin{tabular}{c|c|c|c|c}
\toprule
Selection    & \multirow{2}{*}{Method name} & \multirow{2}{*}{$g$} & \multirow{2}{*}{$E$} & \multirow{2}{*}{$k$} \\
method type & & &  &  \\
\hline
\multirow{3}{*}{Direct} & \oracle   &  \begin{tabular}[c]{@{}c@{}}
    $\{1, 64, 256, $ $ 1024,$ $2048, 4096\}$ (\textsection \ref{sec:framework_analysis:granularity})
\end{tabular}   &  \multirow{6}{*}{\begin{tabular}[c]{@{}c@{}}$\{ 4, 16, $\\ $ (32)^*  \}$\end{tabular} }    &   \multirow{6}{*}{\begin{tabular}[c]{@{}c@{}}$\{4096,$ \\ $(8192)^*\}$\end{tabular}} \\ \cline{2-3}
& \lowRank      &  $\{1 \}$  &      &  \\ \cline{2-3}
& \pkm  \citep{pkm}      &  $\{1 \}$  &        &  \\ \cline{1-3}
\multirow{3}{*}{Indirect} & \hash \citep{hash_layer} &  \begin{tabular}[c]{@{}c@{}}
    $\{1, 64, 256, $ $1024,$ $2048, 4096\}$ (\textsection \ref{sec:framework_analysis:granularity})
\end{tabular}  &      &  \\ \cline{2-3}
& \switch  \citep{switch_transformer}   &  $\{4096\}$  &       &  \\ \cline{2-3}
& \pkmFfn  (\textsection \ref{sec:exp:model})       &  $\{1 \}$  &      &  \\
\bottomrule

\end{tabular}
\caption{All the \generalTerm models used in experiments and analysis in \textsection \ref{sec:framework_analysis} --- $g$ is the number of memory cells grouped in a memory block, $k$ is the active memory cells, and $E$ control the sizes of a memory $d_m = E \cdot (4 \cdot d)$. Some settings(\textbf{*}) are only used for \pkm. }
\label{tab:analysis_experiment}
\end{center}
\end{table*}
\paragraph{\pkmFfn} Since the factorized key table in PKM has little ($<0.3\%$) learnable parameter relative to the value table, we propose an indirect variant called \pkmFfn to match the number of parameters of other models like \hash. This variant has memory block size $g=1$ and the same key-value table as \hash. \pkmFfn has a gate whose $\mathbf{g}(\mathbf{x})$ is the same as the $\mathbf{m}$ from a \pkm and $\mathbf{g}_i = m_i$; and no load-balancing is enforced.

\subsection{Language Modeling}
\paragraph{Pretraining Data}
\label{sec:exp:data}
We pretrain all \generalTerm models on a total of 453GB text with 112B tokens from a union of six English-only datasets, including English subset of CC100 and the five datasets used to pretrain RoBERTa \citep{roberta} --- specifically BookCorpus, \text{English Wikipedia}, \text{CC-News}, \text{OpenWebText}, \text{CC-Stories} (details in Appendix \ref{appendix:setting:data}).
We adopt the same Byte-Pair Encoding as GPT-2 \citep{gpt2} and RoBERTa \citep{roberta} with a vocabulary of 50K subword units. All models are trained for 60B tokens. 
\paragraph{Evaluation settings}
\label{sec:exp:eval}
We evaluate our models' ability to predict the next token in a sequence as measured by perplexity. We report both \textbf{in-domain} and \textbf{out-of-domain} perplexity to indicate generalization ability. For out-of-domain, we use validation data from The Pile \citep{pile}, a public dataset that combines data from 22 diverse sources.
\pdfoutput=1
\section{Analysis Results}
\label{sec:framework_analysis}
In this section, we use the proposed unified view to systematically study the design choice of \generalTerm. Specifically, \textbf{(1)}. we study a wide range of block sizes other than the incidental choice used in existing work and investigate its impact on language modeling perplexity (\textsection \ref{sec:framework_analysis:granularity}). \textbf{(2)}. Both direct and indirect block selection methods lead to lower perplexity than a standard FFN, but which type of method has better FLOPs-Perplexity trade-off and what are the relative efficacy and efficiency of different methods require further study (\textsection\ref{sec:framework_analysis:selection}).

\begin{figure*}[th]
    \centering
     \includegraphics[scale=0.75,trim={3em 1.5em 10em 0.5em},clip]{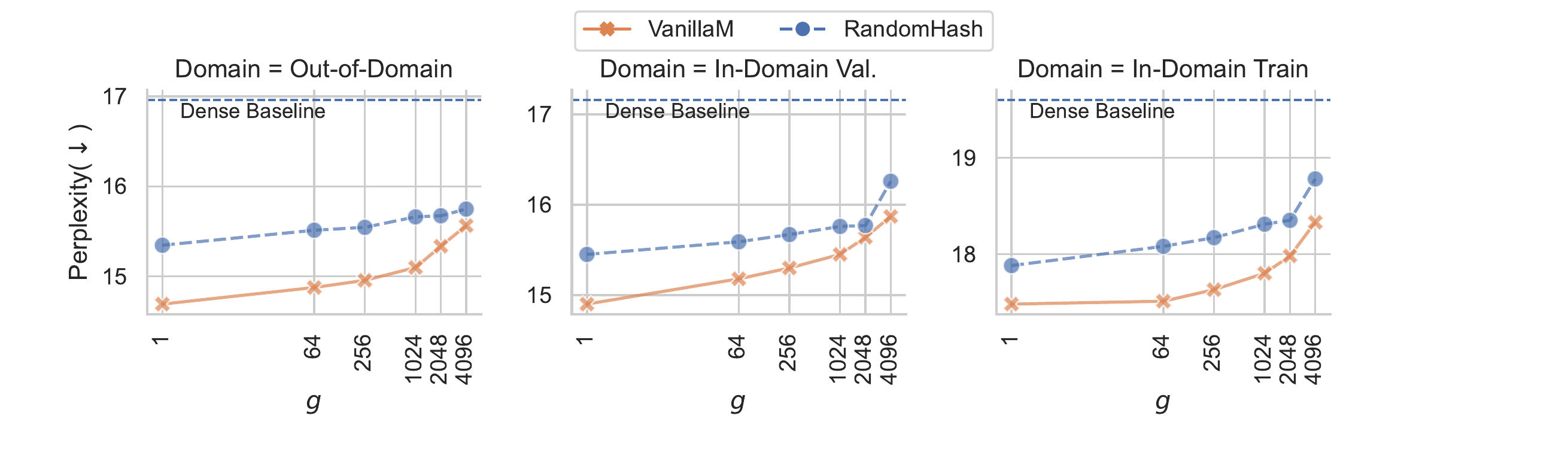}
    \caption{Perplexity (lower the better) consistently improves as memory block size $g$ decreases for both direct (\oracle)  and indirect (\hash) selection method in \generalTerm models. Ranking on individual out-of-domain test sets generally follows the ranking by average perplexity (e.g. 20 out of 22).}
    \label{fig:355m_granularity}
\end{figure*}

\subsection{Memory block size}
\label{sec:framework_analysis:granularity}

Since block size is a natural number, we aim to answer a straightforward question --- given a fixed number of active memory cells $k$, \textbf{does smaller memory block size lead to lower perplexity?} 
We use simple and robust selection methods to disentangle the impact of hyperparameter choices. Specifically, we use random hash as recommended in HashLayer~\citep{hash_layer}  (denoted \textbf{\hash}) for indirect block selection and exact top-$k$ memory block (denoted \textbf{\oracle}) for direct block selection. For all experiments, we use $E = 16$.

\paragraph{\hash} randomly selects $b = k/g $ unique memory blocks among all $B = d_m/g$ blocks --- essentially sampling $b$ unique values from Uniform($[0, \cdots, B-1]$). Originally, with block size $g=4096$, a \hash assigns a token to $4096/4096 = 1$ block; with  block size $g=2048$, $4096 / 2048 = 2$ blocks.

\paragraph{\oracle} originally has a block size $g=1$ and selects top-$k$ scalars in memory coefficients $\mathbf{m} = \text{GeLU}(\mathbf{x} \cdot \mathbf{K}^\top)$. We made a minimal change to extend it to larger block size $g$: given $\mathbf{m}$, we chunk it into $B $ blocks --- $\mathbf{m}^g = [\mathbf{m}^{(0)};\cdots;\mathbf{m}^{(B-1)}]$; then, we select the top-$b$ blocks using the average of each block:
$$\texttt{Avg}(\text{GeLU}(\mathbf{x} \cdot (\mathbf{K}^{(i)})^\top) \texttt{, dim=0})\footnote{\texttt{Avg}$(\cdot)$ performs better than other simple aggregators --- \texttt{Min}$(\cdot)$, \texttt{Max}$(\cdot)$, and \texttt{Avg}$(\mid \cdot \mid)$; see ablations in Table \ref{tab:oracle-aggregator}.}$$

In Fig. \ref{fig:355m_granularity}, we observe that smaller block size leads to an improvement of $0.4 (15.75\rightarrow 15.35)$ perplexity for \hash and an improvement of $0.87(15.56 \rightarrow 14.69)$ for \oracle. 

In Appendix \ref{appendix:granularity:analysis}, we provide theoretical justifications for this observation which shows that a smaller block size improves model capacity by including more combinations of memory cells. For example, with $g/2$, half memory cells of expert-$1$ could be activated together with half of the expert-$2$; however, this combination is impossible with a larger block size.

\pdfoutput=1
\subsection{Memory block selection method}
\label{sec:framework_analysis:selection}
\begin{figure*}[ht]
\centering
 \includegraphics[width=1\linewidth]{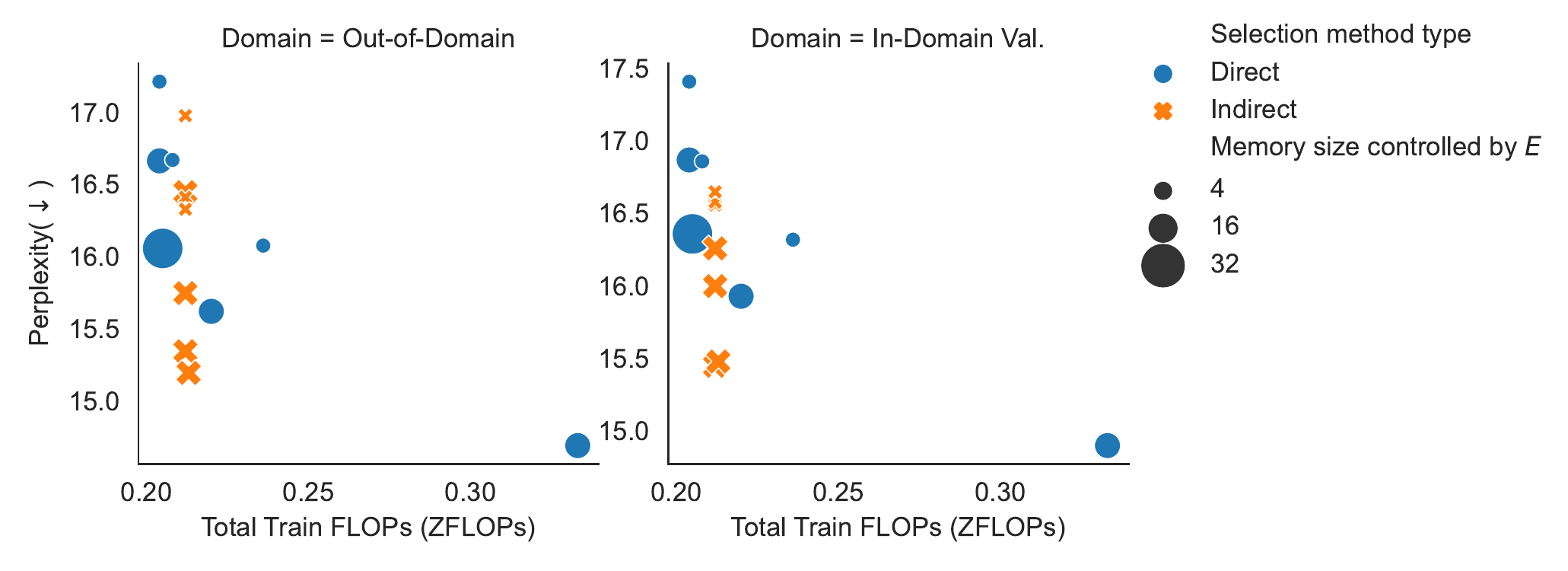}
\caption{FLOPs-perplexity trade-off of indirect block selection is better than that of direct block selection. Indirect methods (orange cross) have more perplexity improvement relative to increases in FLOPs than direct methods (blue dots). See a more detailed legend (e.g. include methods like \lowRank) in Fig. \ref{fig:355m_pareto:detailed}.
}
\label{fig:355m_pareto}
\end{figure*}
Next, we investigate the impact of the selection method, 
specifically, the FLOPs-perplexity trade-off for direct and indirect methods to determine the overall usefulness of each \generalTerm method.


\paragraph{FLOPs-perplexity trade-off} We study the efficiency of direct and indirect selection methods in  \generalTerm models characterized by FLOPS-perplexity trade-off. We conduct experiments across different scales of the memory by varying $E \in \{4, 16\}$; additionally, we run $E=32$ for \pkm. 

In Fig. \ref{fig:355m_pareto}, we marginalize different factors used in the two selection methods --- i.e. types of gates, factorization techniques on key table, etc. --- and consider each type of selection method as a whole. When we change different marginalized factors, we observe that indirect methods tend to improve more as we use more FLOPs (with larger memory sizes controlled by $E$).
Thus, the indirect method has a better FLOPs-perplexity trade-off.

\paragraph{Effect of gating function} We start with contrastive comparisons among \text{\pkmFfn$^{E=16}$}, \text{PKM$^{E=32}$}, \text{RandHash$^{E=16}$} with memory block size $g=1$ and 4096 active memory blocks. From the three parameter-matched models, we can learn important lessons to improve the design of the gate:

\pdfoutput=1
\begin{table*}[ht]
\renewcommand{\arraystretch}{1.6}
\setlength{\tabcolsep}{3.5pt}
\small
\begin{center}

\begin{tabular}{cc|ccc|ccccc}
\toprule
\multirow{3}{*}{\begin{tabular}[c]{@{}c@{}}
    Selection \\
    method type \\
\end{tabular}} & \multirow{3}{*}{Name} &  \multirow{3}{*}{\begin{tabular}[c]{@{}c@{}}
    \# Active \\
    memory \\
    cells $(k)$
\end{tabular}} & \multirow{3}{*}{\begin{tabular}[c]{@{}c@{}}
    \#Parameters \\
    (Entire Model)
\end{tabular}} &  \multirow{3}{*}{\begin{tabular}[c]{@{}c@{}}
    Train \\
    ZFLOPs
\end{tabular}}  &  \multirow{3}{*}{\begin{tabular}[c]{@{}c@{}}
    Out-of-Domain Avg. ($\downarrow$) \\
    (See Table \ref{tab:355m_full_key:OOD} for each domain)
\end{tabular}} \multirow{1}{*}{} &  & \multicolumn{2}{c}{\multirow{2}{*}{In-Domain ($\downarrow$)}} & \\ 
 &   &  &  &   &    & &   &  & \\ \cline{8-9}
 &   &  &  &   &    & &   Train &  Val. & \\
\midrule
& \baseline &  4096 &      354.7M & 0.212 &         16.96       \comment{$\pm$          5.20} &  &          19.60 &           17.16 &  \\ \hline
\multirow{4}{*}{Direct} & PKM$^{E=16}$ & 4096 &      590.2M & 0.205 &      16.66      \comment{$\pm$           5.09} & &          19.45 &           16.87 &   \\ 
 &  PKM$^{E=32}$ & 4096 &      858.7M & 0.205 &        16.06   \comment{ $\pm$             4.84} &  &           18.93 &           16.36 &  \\
 &  PKM$^{E=32}$  & 8192 &      858.7M & 0.213 &        16.16    \comment{$\pm$        4.88} & &            19.05 &           16.45 &  \\ \cline{2-10}
 &  \multirow{1}{*}{VanillaM$^{E=16}$}        & 4096 &      858.3M & 0.333 &     \textbf{14.69}    \comment{$\pm$      \textbf{4.31}} &  &           \textbf{17.48} &           \textbf{14.90}&   \\ \hline
\multirow{2}{*}{Indirect} & \multirow{1}{*}{PKM-FFN$^{E=16}$} & 4096 &      858.9M & 0.213 &     \textbf{15.19}      \comment{$\pm$           \textbf{4.54}} & &            \textbf{17.82}  &           \textbf{15.48} &  \\ \cline{2-10}
 &   \multirow{1}{*}{RandHash$^{E=16}$}     & 4096 &      858.3M & 0.212 &        15.35    \comment{$\pm$          4.59} &  &           17.88 &           15.45 &   \\
\bottomrule 
\end{tabular}
\caption{List of experiments for contrastively comparing designs. This table assume each memory cell is a memory block, i.e. $g=1$. The top two best performing models (bolded) have full-parameter key table and depend more on dot product to activate parameters. Ranking on individual out-of-domain test set generally follows the ranking by average perplexity (e.g. 21 out of 22).}
\label{tab:355m_full_key}
\end{center}
\end{table*}
\begin{enumerate}
    \item Comparing with PKM-FFN$^{E=16}$, PKM$^{E=32}$ essentially moves the parameters from a full-parameter key table to double the size of the value table.
    \item PKM-FFN$^{E=16}$ and \text{RandHash$^{E=16}$} have the same (in size) key and value tables. But the former uses a gate jointly learned with a key table, while the latter uses a learning-free gate.
\end{enumerate}
As shown in Table \ref{tab:355m_full_key}, on out-of-domain, PKM-FFN$^{E=16}$ outperforms \text{PKM$^{E=32}$}(16.06) by $0.87$ perplexity and slightly outperform \text{RandHash$^{E=16}$} by 0.16.
Therefore, it is essential to have \textbf{a full-parameter, and thus expressive enough, key table} to produce memory coefficients.


Table \ref{tab:355m_full_key} shows the improvement of $\oracle^{E=16}$, PKM-FFN$^{E=16}$, \text{RandHash$^{E=16}$} over \baseline (16.96) are 2.27, 1.77, and 1.61 respectively on out-of-domain.  They only differ by how much they depend on the key table for selection --- \oracle directly uses it, \pkmFfn indirectly gains information from it, and \hash completely ignores it.
Due to the consistent observation across out-of-domain test sets, we conclude that \textbf{the more dependent on the key table the selection method is, the better} language model it will lead to; and indirect usage (\pkmFfn) is not enough.

\subsection{Performance of \newMethod}
\label{sec:framework_analysis:avgK}
\pdfoutput=1
\begin{table*}[ht]
\renewcommand{\arraystretch}{1.25}
\setlength{\tabcolsep}{3 pt}
\small
\begin{center}
\begin{tabular}{c|c|ccc|cccc}
\toprule
\multirow{2}{*}{$E$} & \multirow{1}{*}{\#Parameters} & \multirow{2}{*}{Selection method} &  \multirow{2}{*}{$g$} &  \multirow{1}{*}{Train} &  Out-of-Domain Avg. ($\downarrow$) &  \multicolumn{2}{c}{In-Domain  ($\downarrow$)}  &  \\ \cline{7-8}
 & (Entire Model)  &  &  & \multirow{1}{*}{ZFLOPs} &  (See Table \ref{tab:new_method:comparison:OOD} for details) &  Train &  Val. \\
\midrule
1 &   354.7M &  \baseline  &    1 &                           0.212 &          16.96  \comment{$\pm$   5.20} &            19.60 &           17.16 \\ \hline
\multirow{7}{*}{16}  & \multirow{7}{*}{$\approx$ 858.3M} &  \multirow{2}{*}{\begin{tabular}[c]{@{}c@{}}
    \hash  \\
    \citep{hash_layer}
\end{tabular}} & 4096 &                            0.212 &           15.75 \comment{$\pm$ 4.73}   &            18.78 &           16.26 \\\cline{4-8}
 & & & 1 &         0.212 &          15.35   \comment{$\pm$    4.59} &            17.88 &           15.45 \\\cline{3-8}
& &\begin{tabular}[c]{@{}c@{}}
    \switch  \\
    \citep{switch_transformer}
\end{tabular} & 4096 &        0.212 &          16.45  \comment{$\pm$    5.06} &            18.20 &           16.00 \\ \cline{3-8}
 & & \multirow{1}{*}{\pkmFfn} & 1 &         0.213 &          15.19   \comment{$\pm$    4.54} &            17.82 &           15.48 \\\cline{3-8}
& & \multirow{3}{*}{\newMethod} &  4096 &   0.212 &          16.44    \comment{$\pm$  4.98} &            19.04 &           16.59 \\
 & &  &  256 &     0.213 &          14.91   \comment{$\pm$   4.40} &            17.57 &           15.19 \\
 & &  &  64 &      0.214 &          \textbf{14.80   \comment{$\pm$    4.39}} &            \textbf{17.51} &           \textbf{15.11} \\
\bottomrule

\end{tabular}
\caption{\newMethod out-performs other indirect block selection methods. Switch transformer is trained with the load balancing loss to prevent model degradation \citep{first_modern_moe, factored_representation_moe}. Ranking on individual out-of-domain test set mostly follows the ranking by average perplexity (e.g. 21 out of 22).}
\label{tab:new_method:comparison}
\end{center}
\end{table*}

We benchmark our proposed \newMethod approach (\S\ref{sec:new_routing}) in this section.

\paragraph{Language Modeling Pretraining.} Table \ref{tab:new_method:comparison} shows that the proposed \generalTerm design outperforms all other indirect methods.
With $<1\%$ additional FLOPs, \newMethod achieves 2.16 lower perplexity than \baseline(16.96), outperform \citet{switch_transformer} by 1.65 and \citet{hash_layer} by 0.5. %

\paragraph{Analysis} In Fig \ref{fig:new_method:granularity_with_oracle}, although \oracle increases its performance from block size $g=4096$ to $256$, the improvement is less significant than that of \newMethod. The comparison suggests that, with a larger block size, GeLU activation protects the average operation in \oracle (after GeLU) affected by (potentially many) large negatives because $ \lim_{x \to -\infty} \text{GeLU}(x) = 0$. In contrast, this ``negative value'' problem is mitigated by using a smaller block size, due to more blocks available for selection. Since negative dot products affect \newMethod more, it prefers blocks with more or very positive dot products, whereas \oracle is not shielded from extreme negatives, so it might fail to detect those blocks. Therefore, \newMethod could achieve an even slightly better perplexity than \oracle for block size $g \leq 256$. See full discussions in Appendix \ref{appendix:newMethod:analysis}.

In Figure \ref{fig:new_method:load-balance}, we also include a load balancing analysis of \newMethod. To our surprise, the mode collapse (i.e., imbalanced usage of memory block) issue in \newMethod still exists as load balancing loss is not enforced. Given its superior performance, this suggests that \newMethod learned a good representation for each memory block despite the disadvantage of load imbalance. 

\pdfoutput=1
\section{Related Work}
\label{sec:related}

\paragraph{Excluded \generalTerm} Terraformer's \citet{sparse_is_enough} technique on FFN is closest to our \pkmFfn because there is a low-rank learned gate to operate on each memory cells for selection. However, we exclude this method because our framework uses all memory cells in each block, but Terraformer selects $1$ cell in each memory block (see our study in Appendix \ref{appendix:related:terraformer}). In finetuning scenarios, \citet{ffn_are_moe_finetune} studies the connection between FFN and SMoE by turning \textit{trained} FFN into experts and separately learning a gate. In contrast, we focus on pretraining \textit{from scratch}.

\paragraph{Approximate Nearest Neighbour (ANN) search}
One might wonder whether ANN techniques could help to search for the best key in \oracle rather than trade the expressiveness of the key table for efficiency. For example, one could process the unfactorized key table by ANN methods like FAISS \citep{faiss} and ScaNN \citep{scann}. One successful example is applying vanilla Locality-Sensitive Hashing to Reformer \citep{reformer}. However, in our preliminary study, we found that perplexity is greatly affected by the search quality, and building a data structure after every update is expensive and hard to avoid. We leave detailed discussion to Appendix \ref{appendix:related:ann}. 

\pdfoutput=1
\section{Conclusion}
We provide a unified framework for designing sparse FFN in transformers and analyze existing \generalTerm methods such as MoEs in the language modeling task. Using this framework based on sparse neural memory, we found that smaller memory block (e.g. expert) size improves perplexity at the cost of slightly higher computation cost. Selection methods with gates have better FLOPs-Perplexity trade-offs than without, while the gating function in current MoEs is sub-optimal. This framework enables us to instantiate a simpler \generalTerm architecture that outperforms MoEs while still being efficient in training and inference.
\pdfoutput=1
\section*{Limitations} 

\paragraph{Limitations of a smaller block size $g$} With model parallelism \citep{gshard}, multiple GPUs contains different memory block and parallelize the calculations. If with block size $g=4 \cdot d$, a token is only routed to 1 memory block on one device, each device doubles its chance to receive more tokens with block size $g=2 \cdot d$.
Therefore, each GPU processes more tokens and requires more \textit{computation time}, but we didn't measure the wall time difference in our work. Better implementations could be developed to make a smaller block size for practical usage.
In addition, since each memory block has its representation stored in the gating function, the smaller block will lead to more block representation stored in the gate, e.g., more learned parameters in the learned gate and a larger table for the static gate. Although \hash with memory block size $g=4\cdot d$ cost essentially the same with memory block size $g=1$, computing $\mathbf{g}(\mathbf{x})$ for learned gates requires more cost (details in Appendix \ref{appendix:granularity:cost:gate}). 

As discussed in Appendix \ref{appendix:granularity:cost:communication}, smaller memory block size will induce higher communication cost given the current \texttt{all\_to\_all}-based implementation framework (e.g. Switch Transformer). We think reducing memory block size to 1 is too extreme to be practical; and there should be a sweet spot between 1 and 4096 (or the chosen expert size) allowed by the implementation and hardware status.

\paragraph{Limitations of the unified framework}
Since our method \newMethod essentially applies an average pooling to the key table $\mathbf{K}^{g}$, a better alternative may exist. Our method also heavily depends on dot product information, but this might not be the best information to be used. Due to the curse of dimensionality, future work might want to focus on finding a better metric than dot product and other aggregation methods than average to measure distance between high-dimensional vectors. 

Also, we didn't train \newMethod with load-balancing due to our current limit in budget, but we include our rationale in Appendix \ref{appendix:newMethod:load-balancing-rationale} for why \newMethod should work with load-balancing.

Additionally, in large-scale \smoe training, the speed is limited by the most heavy-loaded GPU when model parallelism is used. Therefore, load balancing is essential. We also note that our scale is relatively small and does not use model parallelism, so the problem is not pronounced for us. Future follow-up should look at how to incorporate load balancing into the unified framework and inspire better actionable design choice. We think such unification requires more advanced theoretical connection with memory block and block selection method, which likely involves consideration of training procedure.


\pdfoutput=1

\section*{Ethics Statements}

Due to the nature of pretraining, the carbon footprint of our work is large estimated by the amount of FLOPs and GPUs reported in the paper. We did make the effort to minimize the cost at design stage of the project. In our preliminary study, we ask for recommendation from one of the authors of \citet{fb_moe} to choose and verify the minimal model size and amount of tokens that sufficiently differentiate different design choices.

Another ethical concern of the paper is from the pretraining data we use. As we used the same data source as \citet{fb_moe}, we refer the reader to the ethics statements in  \citet{fb_moe} for how much trained model absorbs bias and toxicity from training data. 
\pdfoutput=1

\section*{Acknowledgement}

We would like to thank (in random order) helpful feedbacks from Suchin Gururangan, Xiaochuang Han, Luke Zettlemoyer, Noah Smith, pre-doctoral members of Noah's Ark, and anonymous reviewers. In developing the experimentation of this work, great help has been received from Jingfei Du, Susan Zhang, and researchers at Meta AI. We would also want to thank Zhaoheng Billy Li for his help in developing analytical formula in Appendix \ref{appendix:granularity:analysis}.





\bibliography{anthology,custom}

\begin{thebibliography}{47}
\expandafter\ifx\csname natexlab\endcsname\relax\def\natexlab#1{#1}\fi

\bibitem[{Artetxe et~al.(2021)Artetxe, Bhosale, Goyal, Mihaylov, Ott, Shleifer,
  Lin, Du, Iyer, Pasunuru, Anantharaman, Li, Chen, Akin, Baines, Martin, Zhou,
  Koura, O'Horo, Wang, Zettlemoyer, Diab, Kozareva, and Stoyanov}]{fb_moe}
Mikel Artetxe, Shruti Bhosale, Naman Goyal, Todor Mihaylov, Myle Ott, Sam
  Shleifer, Xi~Victoria Lin, Jingfei Du, Srinivasan Iyer, Ramakanth Pasunuru,
  Giri Anantharaman, Xian Li, Shuohui Chen, Halil Akin, Mandeep Baines, Louis
  Martin, Xing Zhou, Punit~Singh Koura, Brian O'Horo, Jeff Wang, Luke
  Zettlemoyer, Mona~T. Diab, Zornitsa Kozareva, and Ves Stoyanov. 2021.
\newblock \href {http://arxiv.org/abs/2112.10684} {Efficient large scale
  language modeling with mixtures of experts}.
\newblock \emph{CoRR}, abs/2112.10684.

\bibitem[{Borgeaud et~al.(2022)Borgeaud, Mensch, Hoffmann, Cai, Rutherford,
  Millican, Van Den~Driessche, Lespiau, Damoc, Clark et~al.}]{retro}
Sebastian Borgeaud, Arthur Mensch, Jordan Hoffmann, Trevor Cai, Eliza
  Rutherford, Katie Millican, George~Bm Van Den~Driessche, Jean-Baptiste
  Lespiau, Bogdan Damoc, Aidan Clark, et~al. 2022.
\newblock Improving language models by retrieving from trillions of tokens.
\newblock In \emph{International Conference on Machine Learning}, pages
  2206--2240. PMLR.

\bibitem[{Brown et~al.(2020)Brown, Mann, Ryder, Subbiah, Kaplan, Dhariwal,
  Neelakantan, Shyam, Sastry, Askell, Agarwal, Herbert{-}Voss, Krueger,
  Henighan, Child, Ramesh, Ziegler, Wu, Winter, Hesse, Chen, Sigler, Litwin,
  Gray, Chess, Clark, Berner, McCandlish, Radford, Sutskever, and
  Amodei}]{gpt3}
Tom~B. Brown, Benjamin Mann, Nick Ryder, Melanie Subbiah, Jared Kaplan,
  Prafulla Dhariwal, Arvind Neelakantan, Pranav Shyam, Girish Sastry, Amanda
  Askell, Sandhini Agarwal, Ariel Herbert{-}Voss, Gretchen Krueger, Tom
  Henighan, Rewon Child, Aditya Ramesh, Daniel~M. Ziegler, Jeffrey Wu, Clemens
  Winter, Christopher Hesse, Mark Chen, Eric Sigler, Mateusz Litwin, Scott
  Gray, Benjamin Chess, Jack Clark, Christopher Berner, Sam McCandlish, Alec
  Radford, Ilya Sutskever, and Dario Amodei. 2020.
\newblock \href
  {https://proceedings.neurips.cc/paper/2020/hash/1457c0d6bfcb4967418bfb8ac142f64a-Abstract.html}
  {Language models are few-shot learners}.
\newblock In \emph{Advances in Neural Information Processing Systems 33: Annual
  Conference on Neural Information Processing Systems 2020, NeurIPS 2020,
  December 6-12, 2020, virtual}.

\bibitem[{Chi et~al.(2022)Chi, Dong, Huang, Dai, Ma, Patra, Singhal, Bajaj,
  Song, and Wei}]{x-moe}
Zewen Chi, Li~Dong, Shaohan Huang, Damai Dai, Shuming Ma, Barun Patra, Saksham
  Singhal, Payal Bajaj, Xia Song, and Furu Wei. 2022.
\newblock \href {https://doi.org/10.48550/arXiv.2204.09179} {On the
  representation collapse of sparse mixture of experts}.
\newblock \emph{CoRR}, abs/2204.09179.

\bibitem[{Chowdhery et~al.(2022)Chowdhery, Narang, Devlin, Bosma, Mishra,
  Roberts, Barham, Chung, Sutton, Gehrmann, Schuh, Shi, Tsvyashchenko, Maynez,
  Rao, Barnes, Tay, Shazeer, Prabhakaran, Reif, Du, Hutchinson, Pope, Bradbury,
  Austin, Isard, Gur{-}Ari, Yin, Duke, Levskaya, Ghemawat, Dev, Michalewski,
  Garcia, Misra, Robinson, Fedus, Zhou, Ippolito, Luan, Lim, Zoph, Spiridonov,
  Sepassi, Dohan, Agrawal, Omernick, Dai, Pillai, Pellat, Lewkowycz, Moreira,
  Child, Polozov, Lee, Zhou, Wang, Saeta, Diaz, Firat, Catasta, Wei,
  Meier{-}Hellstern, Eck, Dean, Petrov, and Fiedel}]{palm}
Aakanksha Chowdhery, Sharan Narang, Jacob Devlin, Maarten Bosma, Gaurav Mishra,
  Adam Roberts, Paul Barham, Hyung~Won Chung, Charles Sutton, Sebastian
  Gehrmann, Parker Schuh, Kensen Shi, Sasha Tsvyashchenko, Joshua Maynez,
  Abhishek Rao, Parker Barnes, Yi~Tay, Noam Shazeer, Vinodkumar Prabhakaran,
  Emily Reif, Nan Du, Ben Hutchinson, Reiner Pope, James Bradbury, Jacob
  Austin, Michael Isard, Guy Gur{-}Ari, Pengcheng Yin, Toju Duke, Anselm
  Levskaya, Sanjay Ghemawat, Sunipa Dev, Henryk Michalewski, Xavier Garcia,
  Vedant Misra, Kevin Robinson, Liam Fedus, Denny Zhou, Daphne Ippolito, David
  Luan, Hyeontaek Lim, Barret Zoph, Alexander Spiridonov, Ryan Sepassi, David
  Dohan, Shivani Agrawal, Mark Omernick, Andrew~M. Dai,
  Thanumalayan~Sankaranarayana Pillai, Marie Pellat, Aitor Lewkowycz, Erica
  Moreira, Rewon Child, Oleksandr Polozov, Katherine Lee, Zongwei Zhou, Xuezhi
  Wang, Brennan Saeta, Mark Diaz, Orhan Firat, Michele Catasta, Jason Wei,
  Kathy Meier{-}Hellstern, Douglas Eck, Jeff Dean, Slav Petrov, and Noah
  Fiedel. 2022.
\newblock \href {https://doi.org/10.48550/arXiv.2204.02311} {Palm: Scaling
  language modeling with pathways}.
\newblock \emph{CoRR}, abs/2204.02311.

\bibitem[{Clark et~al.(2022)Clark, de~Las~Casas, Guy, Mensch, Paganini,
  Hoffmann, Damoc, Hechtman, Cai, Borgeaud, van~den Driessche, Rutherford,
  Hennigan, Johnson, Millican, Cassirer, Jones, Buchatskaya, Budden, Sifre,
  Osindero, Vinyals, Rae, Elsen, Kavukcuoglu, and
  Simonyan}]{scaling_law_for_smoe}
Aidan Clark, Diego de~Las~Casas, Aurelia Guy, Arthur Mensch, Michela Paganini,
  Jordan Hoffmann, Bogdan Damoc, Blake~A. Hechtman, Trevor Cai, Sebastian
  Borgeaud, George van~den Driessche, Eliza Rutherford, Tom Hennigan, Matthew
  Johnson, Katie Millican, Albin Cassirer, Chris Jones, Elena Buchatskaya,
  David Budden, Laurent Sifre, Simon Osindero, Oriol Vinyals, Jack~W. Rae,
  Erich Elsen, Koray Kavukcuoglu, and Karen Simonyan. 2022.
\newblock \href {http://arxiv.org/abs/2202.01169} {Unified scaling laws for
  routed language models}.
\newblock \emph{CoRR}, abs/2202.01169.

\bibitem[{Devlin et~al.(2019)Devlin, Chang, Lee, and Toutanova}]{bert}
Jacob Devlin, Ming{-}Wei Chang, Kenton Lee, and Kristina Toutanova. 2019.
\newblock \href {https://doi.org/10.18653/v1/n19-1423} {{BERT:} pre-training of
  deep bidirectional transformers for language understanding}.
\newblock In \emph{Proceedings of the 2019 Conference of the North American
  Chapter of the Association for Computational Linguistics: Human Language
  Technologies, {NAACL-HLT} 2019, Minneapolis, MN, USA, June 2-7, 2019, Volume
  1 (Long and Short Papers)}, pages 4171--4186. Association for Computational
  Linguistics.

\bibitem[{Du et~al.(2021)Du, Huang, Dai, Tong, Lepikhin, Xu, Krikun, Zhou, Yu,
  Firat, Zoph, Fedus, Bosma, Zhou, Wang, Wang, Webster, Pellat, Robinson,
  Meier{-}Hellstern, Duke, Dixon, Zhang, Le, Wu, Chen, and Cui}]{glam}
Nan Du, Yanping Huang, Andrew~M. Dai, Simon Tong, Dmitry Lepikhin, Yuanzhong
  Xu, Maxim Krikun, Yanqi Zhou, Adams~Wei Yu, Orhan Firat, Barret Zoph, Liam
  Fedus, Maarten Bosma, Zongwei Zhou, Tao Wang, Yu~Emma Wang, Kellie Webster,
  Marie Pellat, Kevin Robinson, Kathy Meier{-}Hellstern, Toju Duke, Lucas
  Dixon, Kun Zhang, Quoc~V. Le, Yonghui Wu, Zhifeng Chen, and Claire Cui. 2021.
\newblock \href {http://arxiv.org/abs/2112.06905} {Glam: Efficient scaling of
  language models with mixture-of-experts}.
\newblock \emph{CoRR}, abs/2112.06905.

\bibitem[{Eigen et~al.(2014)Eigen, Ranzato, and
  Sutskever}]{factored_representation_moe}
David Eigen, Marc'Aurelio Ranzato, and Ilya Sutskever. 2014.
\newblock \href {http://arxiv.org/abs/1312.4314} {Learning factored
  representations in a deep mixture of experts}.
\newblock In \emph{2nd International Conference on Learning Representations,
  {ICLR} 2014, Banff, AB, Canada, April 14-16, 2014, Workshop Track
  Proceedings}.

\bibitem[{Fedus et~al.(2022)Fedus, Dean, and Zoph}]{moe_survey_2022}
William Fedus, Jeff Dean, and Barret Zoph. 2022.
\newblock \href {https://doi.org/10.48550/ARXIV.2209.01667} {A review of sparse
  expert models in deep learning}.

\bibitem[{Fedus et~al.(2021)Fedus, Zoph, and Shazeer}]{switch_transformer}
William Fedus, Barret Zoph, and Noam~M. Shazeer. 2021.
\newblock Switch transformers: Scaling to trillion parameter models with simple
  and efficient sparsity.
\newblock \emph{ArXiv}, abs/2101.03961.

\bibitem[{Gao et~al.(2020)Gao, Biderman, Black, Golding, Hoppe, Foster, Phang,
  He, Thite, Nabeshima, Presser, and Leahy}]{pile}
Leo Gao, Stella Biderman, Sid Black, Laurence Golding, Travis Hoppe, Charles
  Foster, Jason Phang, Horace He, Anish Thite, Noa Nabeshima, Shawn Presser,
  and Connor Leahy. 2020.
\newblock The {P}ile: An 800gb dataset of diverse text for language modeling.
\newblock \emph{arXiv preprint arXiv:2101.00027}.

\bibitem[{Geva et~al.(2021)Geva, Schuster, Berant, and Levy}]{key_value_memory}
Mor Geva, Roei Schuster, Jonathan Berant, and Omer Levy. 2021.
\newblock \href {https://doi.org/10.18653/v1/2021.emnlp-main.446} {Transformer
  feed-forward layers are key-value memories}.
\newblock In \emph{Proceedings of the 2021 Conference on Empirical Methods in
  Natural Language Processing, {EMNLP} 2021, Virtual Event / Punta Cana,
  Dominican Republic, 7-11 November, 2021}, pages 5484--5495. Association for
  Computational Linguistics.

\bibitem[{Gokaslan and Cohen(2019)}]{open_web_text}
Aaron Gokaslan and Vanya Cohen. 2019.
\newblock Openwebtext corpus.
\newblock
  \emph{\url{http://web.archive.org/save/http://skylion007.github.io/OpenWebTextCorpus}}.

\bibitem[{Guo et~al.(2020)Guo, Sun, Lindgren, Geng, Simcha, Chern, and
  Kumar}]{scann}
Ruiqi Guo, Philip Sun, Erik Lindgren, Quan Geng, David Simcha, Felix Chern, and
  Sanjiv Kumar. 2020.
\newblock \href {https://arxiv.org/abs/1908.10396} {Accelerating large-scale
  inference with anisotropic vector quantization}.
\newblock In \emph{International Conference on Machine Learning}.

\bibitem[{Gururangan et~al.(2021)Gururangan, Lewis, Holtzman, Smith, and
  Zettlemoyer}]{demix}
Suchin Gururangan, Mike Lewis, Ari Holtzman, Noah~A. Smith, and Luke
  Zettlemoyer. 2021.
\newblock \href {http://arxiv.org/abs/2108.05036} {Demix layers: Disentangling
  domains for modular language modeling}.
\newblock \emph{CoRR}, abs/2108.05036.

\bibitem[{Hoffmann et~al.(2022)Hoffmann, Borgeaud, Mensch, Buchatskaya, Cai,
  Rutherford, de~Las~Casas, Hendricks, Welbl, Clark, Hennigan, Noland,
  Millican, van~den Driessche, Damoc, Guy, Osindero, Simonyan, Elsen, Rae,
  Vinyals, and Sifre}]{palm_big}
Jordan Hoffmann, Sebastian Borgeaud, Arthur Mensch, Elena Buchatskaya, Trevor
  Cai, Eliza Rutherford, Diego de~Las~Casas, Lisa~Anne Hendricks, Johannes
  Welbl, Aidan Clark, Tom Hennigan, Eric Noland, Katie Millican, George van~den
  Driessche, Bogdan Damoc, Aurelia Guy, Simon Osindero, Karen Simonyan, Erich
  Elsen, Jack~W. Rae, Oriol Vinyals, and Laurent Sifre. 2022.
\newblock \href {https://doi.org/10.48550/arXiv.2203.15556} {Training
  compute-optimal large language models}.
\newblock \emph{CoRR}, abs/2203.15556.

\bibitem[{Jacobs et~al.(1991)Jacobs, Jordan, Nowlan, and Hinton}]{original_moe}
Robert~A. Jacobs, Michael~I. Jordan, Steven~J. Nowlan, and Geoffrey~E. Hinton.
  1991.
\newblock Adaptive mixtures of local experts.
\newblock \emph{Neural Computation}, 3:79--87.

\bibitem[{Jaszczur et~al.(2021)Jaszczur, Chowdhery, Mohiuddin, Kaiser,
  Gajewski, Michalewski, and Kanerva}]{sparse_is_enough}
Sebastian Jaszczur, Aakanksha Chowdhery, Afroz Mohiuddin, Lukasz Kaiser,
  Wojciech Gajewski, Henryk Michalewski, and Jonni Kanerva. 2021.
\newblock \href
  {https://proceedings.neurips.cc/paper/2021/hash/51f15efdd170e6043fa02a74882f0470-Abstract.html}
  {Sparse is enough in scaling transformers}.
\newblock In \emph{Advances in Neural Information Processing Systems 34: Annual
  Conference on Neural Information Processing Systems 2021, NeurIPS 2021,
  December 6-14, 2021, virtual}, pages 9895--9907.

\bibitem[{Johnson et~al.(2021)Johnson, Douze, and J{\'{e}}gou}]{faiss}
Jeff Johnson, Matthijs Douze, and Herv{\'{e}} J{\'{e}}gou. 2021.
\newblock \href {https://doi.org/10.1109/TBDATA.2019.2921572} {Billion-scale
  similarity search with gpus}.
\newblock \emph{{IEEE} Trans. Big Data}, 7(3):535--547.

\bibitem[{Kaplan et~al.(2020)Kaplan, McCandlish, Henighan, Brown, Chess, Child,
  Gray, Radford, Wu, and Amodei}]{scaling_law_for_lm}
Jared Kaplan, Sam McCandlish, Tom Henighan, Tom~B. Brown, Benjamin Chess, Rewon
  Child, Scott Gray, Alec Radford, Jeffrey Wu, and Dario Amodei. 2020.
\newblock \href {http://arxiv.org/abs/2001.08361} {Scaling laws for neural
  language models}.
\newblock \emph{CoRR}, abs/2001.08361.

\bibitem[{Kitaev et~al.(2020)Kitaev, Kaiser, and Levskaya}]{reformer}
Nikita Kitaev, Lukasz Kaiser, and Anselm Levskaya. 2020.
\newblock \href {https://openreview.net/forum?id=rkgNKkHtvB} {Reformer: The
  efficient transformer}.
\newblock In \emph{8th International Conference on Learning Representations,
  {ICLR} 2020, Addis Ababa, Ethiopia, April 26-30, 2020}. OpenReview.net.

\bibitem[{Lample et~al.(2019)Lample, Sablayrolles, Ranzato, Denoyer, and
  J{\'{e}}gou}]{pkm}
Guillaume Lample, Alexandre Sablayrolles, Marc'Aurelio Ranzato, Ludovic
  Denoyer, and Herv{\'{e}} J{\'{e}}gou. 2019.
\newblock \href
  {https://proceedings.neurips.cc/paper/2019/hash/9d8df73a3cfbf3c5b47bc9b50f214aff-Abstract.html}
  {Large memory layers with product keys}.
\newblock In \emph{Advances in Neural Information Processing Systems 32: Annual
  Conference on Neural Information Processing Systems 2019, NeurIPS 2019,
  December 8-14, 2019, Vancouver, BC, Canada}, pages 8546--8557.

\bibitem[{Lepikhin et~al.(2021)Lepikhin, Lee, Xu, Chen, Firat, Huang, Krikun,
  Shazeer, and Chen}]{gshard}
Dmitry Lepikhin, HyoukJoong Lee, Yuanzhong Xu, Dehao Chen, Orhan Firat, Yanping
  Huang, Maxim Krikun, Noam~M. Shazeer, and Z.~Chen. 2021.
\newblock Gshard: Scaling giant models with conditional computation and
  automatic sharding.
\newblock \emph{ArXiv}, abs/2006.16668.

\bibitem[{Lewis et~al.(2021)Lewis, Bhosale, Dettmers, Goyal, and
  Zettlemoyer}]{base_layer}
Mike Lewis, Shruti Bhosale, Tim Dettmers, Naman Goyal, and Luke Zettlemoyer.
  2021.
\newblock \href {http://proceedings.mlr.press/v139/lewis21a.html} {{BASE}
  layers: Simplifying training of large, sparse models}.
\newblock In \emph{Proceedings of the 38th International Conference on Machine
  Learning, {ICML} 2021, 18-24 July 2021, Virtual Event}, volume 139 of
  \emph{Proceedings of Machine Learning Research}, pages 6265--6274. {PMLR}.

\bibitem[{Liu et~al.(2019)Liu, Ott, Goyal, Du, Joshi, Chen, Levy, Lewis,
  Zettlemoyer, and Stoyanov}]{roberta}
Yinhan Liu, Myle Ott, Naman Goyal, Jingfei Du, Mandar Joshi, Danqi Chen, Omer
  Levy, Mike Lewis, Luke Zettlemoyer, and Veselin Stoyanov. 2019.
\newblock \href {http://arxiv.org/abs/1907.11692} {{RoBERTa}: {A} {Robustly}
  {Optimized} {BERT} {Pretraining} {Approach}}.

\bibitem[{Nagel(2016)}]{cc_news}
Sebastian Nagel. 2016.
\newblock Cc-news.
\newblock
  \emph{\url{http://web.archive.org/save/http://commoncrawl.org/2016/10/news-dataset-available}}.

\bibitem[{Narayanan et~al.(2021)Narayanan, Shoeybi, Casper, LeGresley, Patwary,
  Korthikanti, Vainbrand, Kashinkunti, Bernauer, Catanzaro, Phanishayee, and
  Zaharia}]{large_scale_with_megatron_lm}
Deepak Narayanan, Mohammad Shoeybi, Jared Casper, Patrick LeGresley, Mostofa
  Patwary, Vijay Korthikanti, Dmitri Vainbrand, Prethvi Kashinkunti, Julie
  Bernauer, Bryan Catanzaro, Amar Phanishayee, and Matei Zaharia. 2021.
\newblock \href {https://doi.org/10.1145/3458817.3476209} {Efficient
  large-scale language model training on {GPU} clusters using megatron-lm}.
\newblock In \emph{{SC} '21: The International Conference for High Performance
  Computing, Networking, Storage and Analysis, St. Louis, Missouri, USA,
  November 14 - 19, 2021}, pages 58:1--58:15. {ACM}.

\bibitem[{Paszke et~al.(2019)Paszke, Gross, Massa, Lerer, Bradbury, Chanan,
  Killeen, Lin, Gimelshein, Antiga, Desmaison, K{\"{o}}pf, Yang, DeVito,
  Raison, Tejani, Chilamkurthy, Steiner, Fang, Bai, and Chintala}]{pytorch}
Adam Paszke, Sam Gross, Francisco Massa, Adam Lerer, James Bradbury, Gregory
  Chanan, Trevor Killeen, Zeming Lin, Natalia Gimelshein, Luca Antiga, Alban
  Desmaison, Andreas K{\"{o}}pf, Edward~Z. Yang, Zachary DeVito, Martin Raison,
  Alykhan Tejani, Sasank Chilamkurthy, Benoit Steiner, Lu~Fang, Junjie Bai, and
  Soumith Chintala. 2019.
\newblock \href
  {https://proceedings.neurips.cc/paper/2019/hash/bdbca288fee7f92f2bfa9f7012727740-Abstract.html}
  {Pytorch: An imperative style, high-performance deep learning library}.
\newblock In \emph{Advances in Neural Information Processing Systems 32: Annual
  Conference on Neural Information Processing Systems 2019, NeurIPS 2019,
  December 8-14, 2019, Vancouver, BC, Canada}, pages 8024--8035.

\bibitem[{Radford and Narasimhan(2018)}]{gpt1}
Alec Radford and Karthik Narasimhan. 2018.
\newblock Improving language understanding by generative pre-training.

\bibitem[{Radford et~al.(2019)Radford, Wu, Child, Luan, Amodei, and
  Sutskever}]{gpt2}
Alec Radford, Jeff Wu, Rewon Child, David Luan, Dario Amodei, and Ilya
  Sutskever. 2019.
\newblock Language models are unsupervised multitask learners.

\bibitem[{Raffel et~al.(2020)Raffel, Shazeer, Roberts, Lee, Narang, Matena,
  Zhou, Li, and Liu}]{t5}
Colin Raffel, Noam Shazeer, Adam Roberts, Katherine Lee, Sharan Narang, Michael
  Matena, Yanqi Zhou, Wei Li, and Peter~J. Liu. 2020.
\newblock \href {http://jmlr.org/papers/v21/20-074.html} {Exploring the limits
  of transfer learning with a unified text-to-text transformer}.
\newblock \emph{J. Mach. Learn. Res.}, 21:140:1--140:67.

\bibitem[{Roller et~al.(2021)Roller, Sukhbaatar, Szlam, and
  Weston}]{hash_layer}
Stephen Roller, Sainbayar Sukhbaatar, Arthur Szlam, and Jason Weston. 2021.
\newblock \href {http://arxiv.org/abs/2106.04426} {Hash layers for large sparse
  models}.
\newblock \emph{CoRR}, abs/2106.04426.

\bibitem[{Schwartz et~al.(2020)Schwartz, Dodge, Smith, and Etzioni}]{green_ai}
Roy Schwartz, Jesse Dodge, Noah Smith, and Oren Etzioni. 2020.
\newblock Green ai.
\newblock \emph{Communications of the ACM}, 63:54 -- 63.

\bibitem[{Shazeer et~al.(2017)Shazeer, Mirhoseini, Maziarz, Davis, Le, Hinton,
  and Dean}]{first_modern_moe}
Noam Shazeer, Azalia Mirhoseini, Krzysztof Maziarz, Andy Davis, Quoc~V. Le,
  Geoffrey~E. Hinton, and Jeff Dean. 2017.
\newblock \href {https://openreview.net/forum?id=B1ckMDqlg} {Outrageously large
  neural networks: The sparsely-gated mixture-of-experts layer}.
\newblock In \emph{5th International Conference on Learning Representations,
  {ICLR} 2017, Toulon, France, April 24-26, 2017, Conference Track
  Proceedings}. OpenReview.net.

\bibitem[{Sukhbaatar et~al.(2019)Sukhbaatar, Grave, Lample, Jegou, and
  Joulin}]{sukhbaatar2019augmenting}
Sainbayar Sukhbaatar, Edouard Grave, Guillaume Lample, Herve Jegou, and Armand
  Joulin. 2019.
\newblock Augmenting self-attention with persistent memory.
\newblock \emph{arXiv preprint arXiv:1907.01470}.

\bibitem[{Sukhbaatar et~al.(2015{\natexlab{a}})Sukhbaatar, Szlam, Weston, and
  Fergus}]{end2end_memory_net}
Sainbayar Sukhbaatar, Arthur~D. Szlam, Jason Weston, and Rob Fergus.
  2015{\natexlab{a}}.
\newblock End-to-end memory networks.
\newblock In \emph{NIPS}.

\bibitem[{Sukhbaatar et~al.(2015{\natexlab{b}})Sukhbaatar, Weston, Fergus
  et~al.}]{sukhbaatar2015end}
Sainbayar Sukhbaatar, Jason Weston, Rob Fergus, et~al. 2015{\natexlab{b}}.
\newblock End-to-end memory networks.
\newblock \emph{Advances in neural information processing systems}, 28.

\bibitem[{Tay et~al.(2020)Tay, Dehghani, Bahri, and Metzler}]{tay2020efficient}
Yi~Tay, Mostafa Dehghani, Dara Bahri, and Donald Metzler. 2020.
\newblock Efficient transformers: A survey.
\newblock \emph{ACM Computing Surveys (CSUR)}.

\bibitem[{Trinh and Le(2018)}]{cc_stories}
Trieu~H. Trinh and Quoc~V. Le. 2018.
\newblock A simple method for commonsense reasoning.
\newblock \emph{ArXiv}, abs/1806.02847.

\bibitem[{Vaswani et~al.(2017)Vaswani, Shazeer, Parmar, Uszkoreit, Jones,
  Gomez, Kaiser, and Polosukhin}]{vanilla_transformer}
Ashish Vaswani, Noam~M. Shazeer, Niki Parmar, Jakob Uszkoreit, Llion Jones,
  Aidan~N. Gomez, Lukasz Kaiser, and Illia Polosukhin. 2017.
\newblock Attention is all you need.
\newblock \emph{ArXiv}, abs/1706.03762.

\bibitem[{Wei et~al.(2022)Wei, Tay, Bommasani, Raffel, Zoph, Borgeaud,
  Yogatama, Bosma, Zhou, Metzler, Chi, Hashimoto, Vinyals, Liang, Dean, and
  Fedus}]{emergent_ability_of_large_language_models}
Jason Wei, Yi~Tay, Rishi Bommasani, Colin Raffel, Barret Zoph, Sebastian
  Borgeaud, Dani Yogatama, Maarten Bosma, Denny Zhou, Donald Metzler, Ed~H.
  Chi, Tatsunori Hashimoto, Oriol Vinyals, Percy Liang, Jeff Dean, and William
  Fedus. 2022.
\newblock \href {https://doi.org/10.48550/arXiv.2206.07682} {Emergent abilities
  of large language models}.
\newblock \emph{CoRR}, abs/2206.07682.

\bibitem[{Wenzek et~al.(2020)Wenzek, Lachaux, Conneau, Chaudhary, Guzm'an,
  Joulin, and Grave}]{cc_net}
Guillaume Wenzek, Marie-Anne Lachaux, Alexis Conneau, Vishrav Chaudhary,
  Francisco Guzm'an, Armand Joulin, and Edouard Grave. 2020.
\newblock Ccnet: Extracting high quality monolingual datasets from web crawl
  data.
\newblock In \emph{LREC}.

\bibitem[{Zhang et~al.(2022{\natexlab{a}})Zhang, Roller, Goyal, Artetxe, Chen,
  Chen, Dewan, Diab, Li, Lin, Mihaylov, Ott, Shleifer, Shuster, Simig, Koura,
  Sridhar, Wang, and Zettlemoyer}]{opt}
Susan Zhang, Stephen Roller, Naman Goyal, Mikel Artetxe, Moya Chen, Shuohui
  Chen, Christopher Dewan, Mona Diab, Xian Li, Xi~Victoria Lin, Todor Mihaylov,
  Myle Ott, Sam Shleifer, Kurt Shuster, Daniel Simig, Punit~Singh Koura, Anjali
  Sridhar, Tianlu Wang, and Luke Zettlemoyer. 2022{\natexlab{a}}.
\newblock \href {https://doi.org/10.48550/arXiv.2205.01068} {{OPT:} open
  pre-trained transformer language models}.
\newblock \emph{CoRR}, abs/2205.01068.

\bibitem[{Zhang et~al.(2022{\natexlab{b}})Zhang, Lin, Liu, Li, Sun, and
  Zhou}]{ffn_are_moe_finetune}
Zhengyan Zhang, Yankai Lin, Zhiyuan Liu, Peng Li, Maosong Sun, and Jie Zhou.
  2022{\natexlab{b}}.
\newblock \href {https://doi.org/10.18653/v1/2022.findings-acl.71}
  {{M}o{E}fication: Transformer feed-forward layers are mixtures of experts}.
\newblock In \emph{Findings of the Association for Computational Linguistics:
  ACL 2022}, pages 877--890, Dublin, Ireland. Association for Computational
  Linguistics.

\bibitem[{Zhou et~al.(2022)Zhou, Lei, Liu, Du, Huang, Zhao, Dai, Chen, Le, and
  Laudon}]{moe_by_topk_token}
Yanqi Zhou, Tao Lei, Hanxiao Liu, Nan Du, Yanping Huang, Vincent~Y. Zhao,
  Andrew~M. Dai, Zhifeng Chen, Quoc Le, and James Laudon. 2022.
\newblock \href {http://arxiv.org/abs/2202.09368} {Mixture-of-experts with
  expert choice routing}.
\newblock \emph{CoRR}, abs/2202.09368.

\bibitem[{Zhu et~al.(2015)Zhu, Kiros, Zemel, Salakhutdinov, Urtasun, Torralba,
  and Fidler}]{bookcorpus}
Yukun Zhu, Ryan Kiros, Richard~S. Zemel, Ruslan Salakhutdinov, Raquel Urtasun,
  Antonio Torralba, and Sanja Fidler. 2015.
\newblock Aligning books and movies: Towards story-like visual explanations by
  watching movies and reading books.
\newblock \emph{2015 IEEE International Conference on Computer Vision (ICCV)},
  pages 19--27.

\end{thebibliography}
\bibliographystyle{acl_natbib}

\appendix

\pdfoutput=1
\section{Experimental details}
\label{appendix:setting}

\subsection{Hyperparameters}
\label{appendix:setting:hyper}
Table \ref{tab:hyper:shared} specifies shared hyperparameters across all experiments, in which Table \ref{tab:hyper:shared:data+optim+infra} contains ones for training data, optimizer, and efficient infrastructure techniques; and Table \ref{tab:hyper:shared:arch} for architecture. Then, Table \ref{tab:hyper:specific:switch} describes the hyperparameters specifically for \switch, Table \ref{tab:hyper:specific:lowrank} for \lowRank,  Table \ref{tab:hyper:specific:pkm} for \pkm. In preliminary study, we train baselines with different random seeds and found the results are almost identical. Therefore, in the interest of time and carbon cost, we didn't run with different random seeds for the experiments in this paper.  
\pdfoutput=1
\begin{table*}[h] 
    \centering
    \begin{subtable}[h]{1\linewidth}
        \centering
        \caption{Shared configuration for data, optimizer, and efficient infrastructure}
        \begin{tabular}{ll}
            \toprule
            \textbf{Name} & \textbf{Values}  \\
            \toprule
                \#Tokens for training & 60e9 \\
                \#Tokens for warm-up & $375 \cdot 1024^2$ \\
                \#Tokens per batch & $0.5 \cdot 1024^2$ \\
                \#Tokens per sample & 2048 \\
                \#GPU & 32 \\
                GPU & NVIDIA Tesla V100 32GB \\
                Optimizer & Adam$(\beta$s $= (0.9, 0.98)$, $\epsilon=1e-8)$ \\
                Weight Decay & 0.01 \\
                Peak Learning Rate & 3e-4 \\
                Learning Rate Scheduler & \texttt{polynomial\_decay} \\
                \texttt{clip\_norm} & 0.0\\
                DistributedDataParallel backend & FullyShardedDataParallel\\
                \texttt{memory-efficient-fp16} & True \\
                \texttt{fp16-init-scale} & 4 \\
                \texttt{checkpoint-activations} & True \\
            \bottomrule
        \end{tabular}
        \label{tab:hyper:shared:data+optim+infra}
    \end{subtable} 
    \\
    \begin{subtable}[h]{1\linewidth}
        \centering
        \caption{Shared configuration for architecture.}
        \begin{tabular}{ll}
            \toprule
            \textbf{Name} & \textbf{Values}  \\
            \toprule
                Objective & Causal Language Model (CLM)  \\
                Activation function($f$) & GeLU \\
                Model dimension ($d$) & 1024 \\
                $d_m$ of non-\generalTerm & $4 \cdot 1024$ \\
                \#Attention Head & 16 \\
                \#Layer & 24 \\
                Dropout Rate & 0.0 \\
                Attention Dropout Rate & 0.0\\
                \texttt{share-decoder-input-output-embed} & True \\
            \bottomrule
        \end{tabular}
        \label{tab:hyper:shared:arch}
    \end{subtable}
    \caption{Shared configuration. This is also used for training base model.}
    \label{tab:hyper:shared}
\end{table*}

\pdfoutput=1
\begin{table*}[ht!] 
    \centering
    \renewcommand{\arraystretch}{1.2}
    \setlength{\tabcolsep}{4pt}
    \begin{subtable}[ht]{1.\linewidth}
        \centering
        \caption{\switch}
        \label{tab:hyper:specific:switch}
        \begin{tabular}{lc}
            \toprule
            \textbf{Name} & \textbf{Values}  \\
            \toprule
                \texttt{moe-gating-use-fp32} & True \\
                \multirow{2}{*}{\texttt{moe-gate-loss-wt}} & 0.01 \\
                & i.e. CLM loss + $0.01 \cdot $ auxiliary loss~\citep{switch_transformer} \\
                Divide expert gradients by & $\sqrt{\text{\# Expert}} = \sqrt{B}$ \\
            \bottomrule
        \end{tabular}
    \end{subtable}
    \\
    \begin{subtable}[ht]{0.45\linewidth}
        \centering
        \caption{\lowRank}
        \label{tab:hyper:specific:lowrank}
        \begin{tabular}{cc}
            \toprule
            \textbf{Name} & \textbf{Values}  \\
            \toprule
                $d_\ell$ & 128 \\
                BatchNorm after $\mathbf{x} \cdot \mathbf{D}$ & False \\
            \bottomrule
        \end{tabular}
    \end{subtable} 
    \hfill
    \begin{subtable}[ht]{0.45\linewidth}
        \centering
        \caption{\pkm}
        \label{tab:hyper:specific:pkm}
        \begin{tabular}{cc}
            \toprule
            \textbf{Name} & \textbf{Values}  \\
            \toprule
                $d_\ell$ & 128 \\
                \# key table (\textsection \ref{sec:bgd:non-moe}) & 1 \\
                BatchNorm after $\mathbf{x} \cdot \mathbf{D}$ & True \\
            \bottomrule
        \end{tabular}
    \end{subtable}
    \caption{Specific architecture configuration}
    \label{tab:hyper:specific}
\end{table*}


\subsection{Infrastructure}
We used \texttt{fairseq} as our codebase to conduct experiments. For training, all the training are done on V100 GPUs. Each training takes 32 GPUs; in rare circumstances, the number of GPU is adjusted for training speed.

\subsection{Data}
\label{appendix:setting:data}
Here is a detailed description of our pretraining corpus.
\begin{itemize}
    \item \textbf{BookCorpus} \citep{bookcorpus} consists of more than 10K unpublished books (4GB);
    \item \textbf{English Wikipedia}, excluding lists, tables and headers (12GB);
    \item \textbf{CC-News} \citep{cc_news} contains 63 millions English news articles crawled between September
2016 and February 2019 (76GB);
    \item \textbf{OpenWebText} \citep{open_web_text}, an open source recreation of the WebText dataset used to train GPT-2 (38GB);
    \item \textbf{CC-Stories} \citep{cc_stories} contains a subset of CommonCrawl data filtered to match the story-like style of Winograd schemas (31GB);
    \item \textbf{English CC100} \citep{cc_net}, a dataset extracted from CommonCrawl snapshots between January 2018 and December 2018, filtered to match the style of Wikipedia (292GB).
\end{itemize}
\section{Product Key Memory}
\label{appendix:pkm}

As mentioned in \S\ref{sec:bgd:non-moe}, \lowRank assumes that the full key table $\mathbf{K}^\top \in \mathbb{R}^{d\times d_m}$ is composed of and approximated by a downward projection $\mathbf{D} \in \mathbb{R}^{d \times d_\ell}$ and a low-rank key table $\tilde{\mathbf{K}} \in \mathbb{R}^{d_m \times d_\ell}$, 
\begin{equation*}
    \mathbf{K}^\top = \mathbf{D}  \cdot  \tilde{\mathbf{K}}^\top
\end{equation*}

\citet{pkm} further approximated the low-rank key table $\tilde{\mathbf{K}}$. It assumes that each low-rank key vector $\tilde{\mathbf{k}}_i \in \mathbb{R}^{d_\ell}$ is created from the concatenation from two sub-key vectors $\mathbf{c}, \mathbf{c}' \in \mathbb{R}^{\frac{d_\ell}{2}}$; and the two sub-key vectors is from two smaller and non-overlapped \textit{sub-key} tables $\mathbf{C}, \mathbf{C}' \in \mathbb{R}^{\sqrt{d_m} \times \frac{d_\ell}{2}}$. 

Unlike \lowRank where key vectors are independent of each other, key vectors in \pkm have some overlaps with each other and have a \textit{structured sharing},
\begin{equation*}
    \tilde{\mathbf{k}}_i = \left[\;\mathbf{c}_{\floor{i / \sqrt{d_m}}}, \;\; \mathbf{c}'_{i\Mod{\sqrt{d_m}}}\;\right] \in \mathbb{R}^{d_\ell}
\end{equation*}
One could exploit this structure to efficiently compute the memory coefficient $\mathbf{m}$. 
One first calculates the dot product between sub-key tables and down-projected input $\mathbf{t}$ individually and combines them with a negligible cost to form the full dot product $\mathbf{m} \in \mathbb{R}^{d_m}$:
\begin{align*}
    m_i & = f\left(\mathbf{s}_{\floor{i / \sqrt{d_m}}} + \mathbf{s}'_{i\Mod{\sqrt{d_m}}}\right) \\
    \text{ where } &  \mathbf{s} = \mathbf{t}\left[:\frac{d_\ell}{2}\right] \cdot \mathbf{C}^\top \in \mathbb{R}^{\sqrt{d_m}}, \\
    & \mathbf{s}' = \mathbf{t}\left[\frac{d_\ell}{2}:\right] \cdot (\mathbf{C}')^\top \in \mathbb{R}^{\sqrt{d_m}}\\
    & \mathbf{t} = \mathbf{x} \cdot \mathbf{D} \in \mathbb{R}^{\sqrt{d_\ell}}
\end{align*}
\pdfoutput=1
\section{Block size}
\label{appendix:granularity}
\subsection{\oracle with block size $g > 1$}
\label{appendix:granularity:aggregator}
For \oracle with block size $g > 1$, we also tried three other simple aggregation function, but they all under-perform Average. We show their results in Table \ref{tab:oracle-aggregator}.
\pdfoutput=1
\begin{table*}[ht]
\renewcommand{\arraystretch}{1.5}
\setlength{\tabcolsep}{4pt}
\small
\begin{center}

\begin{tabular}{cccc|cccc}
\toprule
 \multirow{2}{*}{Selection method} &   \multirow{2}{*}{$g$} & \#Parameters  &  Train & \multirow{2}{*}{Aggregator} &  Out-of-Domain &    \multicolumn{2}{c}{In-Domain}    \\ \cline{7-8}
 &     & (Entire Model) &  ZFLOPs &    & (22 domains; Avg. $\pm$ Std.) &    Train &   Val. \\
\midrule
Dense Baseline &    1 &      354.7M &           0.212 &             N/A &     16.96  $\pm$ 5.20 &            19.60 &           17.16 \\\hline
\multirow{4}{*}{\oracle} &    \multirow{4}{*}{4096} &      \multirow{4}{*}{858.3M} &                         \multirow{4}{*}{0.333} &    Avg($ \cdot $) &      \textbf{15.56}  $\pm$  \textbf{4.62} &            18.33 &           \textbf{15.87} \\  \cline{5-8}
&   &       &       &    Avg($|\cdot|$) &        15.67    $\pm$    4.66 &            \update{} &           15.94 \\  \cline{5-8}
&   &     &          &     Max($\cdot$) &     16.11  $\pm$  4.86 &            \update{} &           16.33 \\  \cline{5-8}
&   &    &  &   Min($\cdot$) &    94.86 $\pm$  57.63 &            \update{} &           17.08 \\
\bottomrule
\end{tabular}
\caption{\oracle with different simple aggregators}
\label{tab:oracle-aggregator}
\end{center}
\end{table*}
\subsection{Analysis of smaller block sizes}
\label{appendix:granularity:analysis}
We first quantify the intuition ---``usage of model memory is more spread out" by number of activated memory cells shared between two random tokens --- $\mathbb{E}[r]$. We define this quantity to be average of every \generalTerm layer, to reflect the overall behavior 
$$\mathbb{E}[r] = \frac{1}{L_\text{\generalTerm}} \sum_{\ell} \mathbb{E}[r_\ell]$$
where $L_\text{\generalTerm}$ is the number of \generalTerm. Because block selection usually depends on a contextualized token embedding, it's hard to draw tokens in an i.i.d. fashion. Therefore, we estimate the the quantity by evaluating the model on a validation set. We sample $N$ token pairs from each sequence for estimation:
\begin{align*}
    \mathbb{E}[r_\ell] = &  \frac{1}{|\text{val}| \cdot N} \sum_{s \in \text{val}} \\
    & \sum^{N-1}_{i=0: (x, y)_i \sim \text{Uniform}(s \times s)} | \mathcal{I}_x \cap \mathcal{I}_y| \cdot g 
\end{align*}
where $\mathcal{I}_x$ is the indices of selected memory block for token at position $x$, and similarly for $\mathcal{I}_y$.

\hash, though, is an exception where uniform sampling is used. Therefore, $\mathbb{E}[r]$ could also be analytically calculated for various $g$, \textit{when assuming tokens are also uniformly distributed}.\footnote{In the formulae, we use $\cdot$ and $\times$ interchangeably for better presentation}

\begin{align*}
\mathbb{E}[r] =  & \frac{1}{L_\text{\generalTerm}} \cdot L_\text{\generalTerm} \cdot  \mathbb{E}[r_\ell] \\
= \sum_{i=1}^b & \underbrace{{b \choose i}}_\text{\begin{tabular}[c]{@{}c@{}}No. of such \\ block assignments\end{tabular}} \times \\
& \underbrace{\prod_{j=0}^{i-1} \frac{b-j}{B-j}}_\text{\begin{tabular}[c]{@{}c@{}}Probability of \\ $i$ overlaps \end{tabular}} \cdot \underbrace{\prod_{k=0}^{b-i-1} \frac{B-b-k}{B-k}}_\text{\begin{tabular}[c]{@{}c@{}}Probability of \\ $j$ non-overlaps \end{tabular}} \times \\
& \underbrace{i \cdot g}_\text{\begin{tabular}[c]{@{}c@{}} $r$ cells \\ in an overlap \end{tabular}}
\end{align*}

\begin{figure*}[th!]
     \centering
          \begin{subfigure}{1.\linewidth}
         \centering
         \includegraphics[width=1\linewidth]{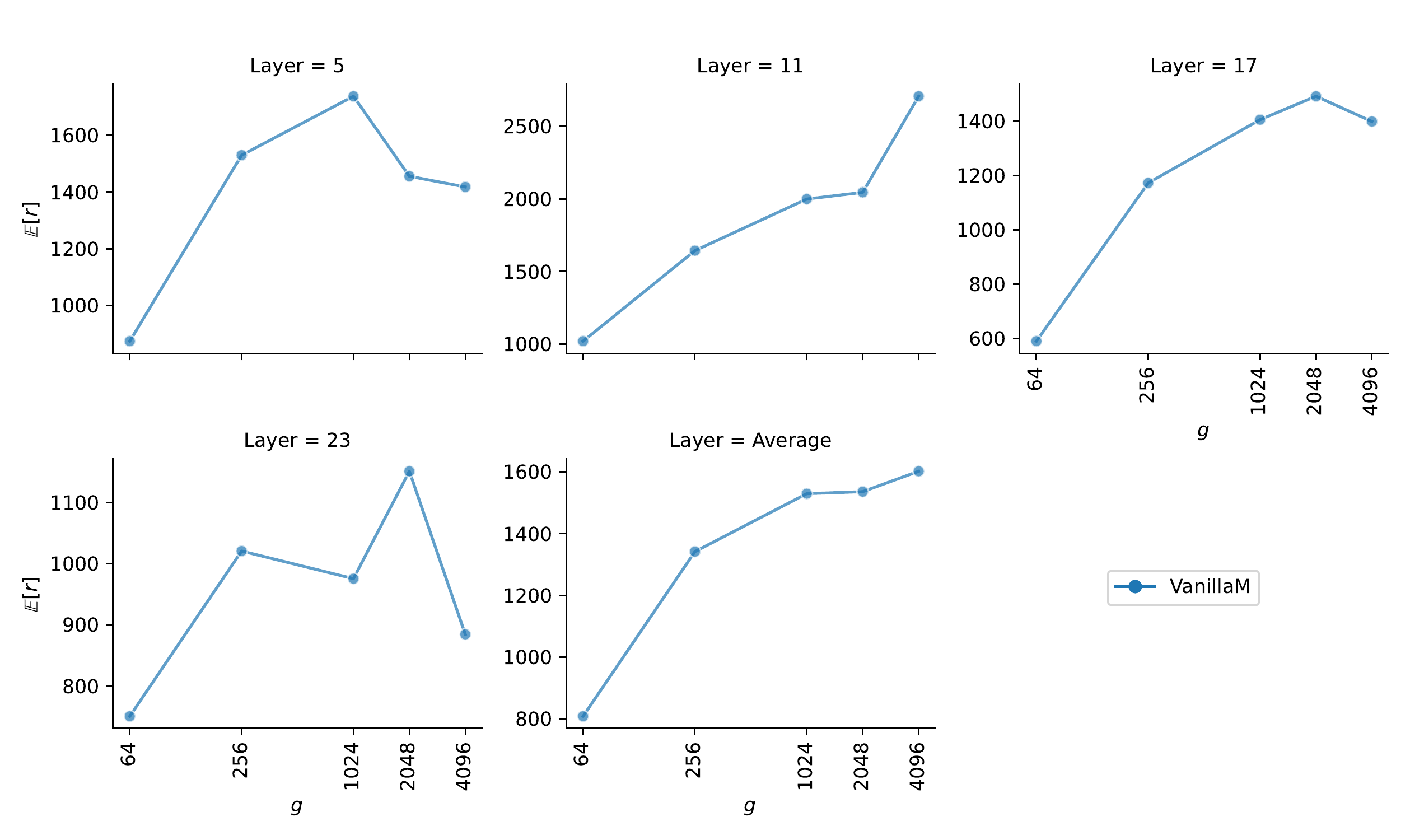}
         \caption{\oracle: empirical estimation from sampling}
         \label{fig:E_r:oracle}
     \end{subfigure}
     \\
     \begin{subfigure}{1\linewidth}
         \centering
         \includegraphics[width=1\linewidth]{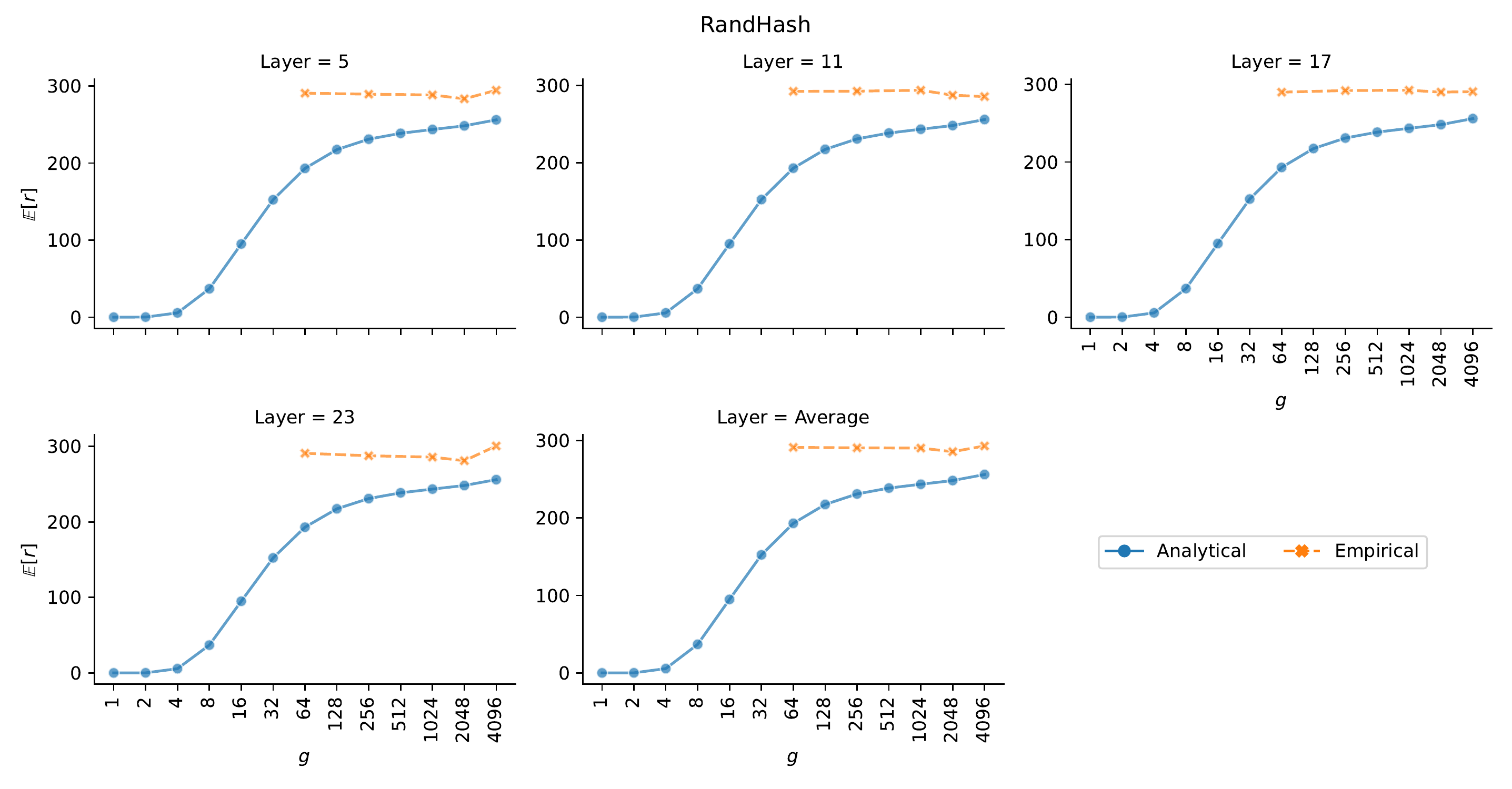}
         \caption{\hash: both analytical value from close form calculation and empirical value from sampling}
         \label{fig:E_r:hash}
     \end{subfigure}

    \caption{Expected of shared memory cells across various block size $g$}
    \label{fig:E_r}
\end{figure*}

In Fig. \ref{fig:E_r:oracle}, \ref{fig:E_r:hash}, we evaluate our model with $E=16$ on our validation subset and calculate the estimations across various $g$.\textbf{ It is observed that less sharing happens as block size decreases.} However, the empirical estimation for \hash are relatively constant across granularity. We suspect this is due to the Zipf's law of tokens. Also, we note that the magnitude of $\mathbb{E}[r]$ are different for different methods. We defer the reason of this phenomena to future work.

\subsection{Cost of smaller block sizes}
\label{appendix:granularity:cost}
\subsubsection{Cost of gate}
\label{appendix:granularity:cost:gate}
\hash is efficient for computation because a hash table theoretically has time complexity $O(1)$. In contrast, a conventional learned gate \ref{sec:bgd:moe} has an $d$-dimensional embedding for each memory block. Therefore, with total of $B$ memory blocks, it has the time complexity of $O(d \cdot B)$. In Table \ref{tab:granuarity:cost} we show how the FLOPs percentage of learned gate in a single forward-backward computation changes with respect to the change in memory block size, where we assume setup in \textsection \ref{sec:exp} is adopted.
\pdfoutput=1
\begin{table*}[ht]
\renewcommand{\arraystretch}{1.5}
\setlength{\tabcolsep}{6pt}
\centering

\begin{tabular}{c|ccccccccc}
\toprule
\multirow{2}{*}{TFLOPs of } & \multicolumn{9}{c}{Memory block size ($g$)}  \\\cline{2-10}
                            & 4096              & 2048     & 1024     & 512      & 256      & 128      & 64       & 32       & 1        \\\hline
\begin{tabular}[c]{@{}c@{}}
    4 learned gates\\
    (across 24 layers)
\end{tabular} & 0.275          & 0.552 & 1.10  & 2.20  & 4.40   & 8.80   & 17.6  & 35.2  & 1124 \\\hline
\begin{tabular}[c]{@{}c@{}}
    Entire model
\end{tabular}      & 1850          & 1850 & 1860 & 1860 & 1860 & 1860 & 1870 & 1890 & 2980 \\\hline
\%       & 0.0149          & 0.0298 & 0.0595 & 0.118  & 0.237  & 0.473  & 0.941  & 1.86   & 37.718 \\\bottomrule
\end{tabular}
\caption{FLOPs percentage of learned gate increases when memory block size $g$ decreases}
\label{tab:granuarity:cost}
\end{table*}

\subsubsection{Cost of communication}
\label{appendix:granularity:cost:communication}
The conventional training framework of MoE \citep{switch_transformer} depends on \texttt{all\_to\_all} operations \citep{pytorch} to route tokens to different devices. One might expect the communication cost remains the same if the number of device doesn't change. However, this assumes the tokens are identified by their type. In fact, the training framework further identify the routed tokens type by the experts it routed to. Therefore, the communication cost scales linearly with respect to the change in the number of memory block.

\pdfoutput=1
\begin{figure*}[ht]
\centering
 \includegraphics[width=1\linewidth]{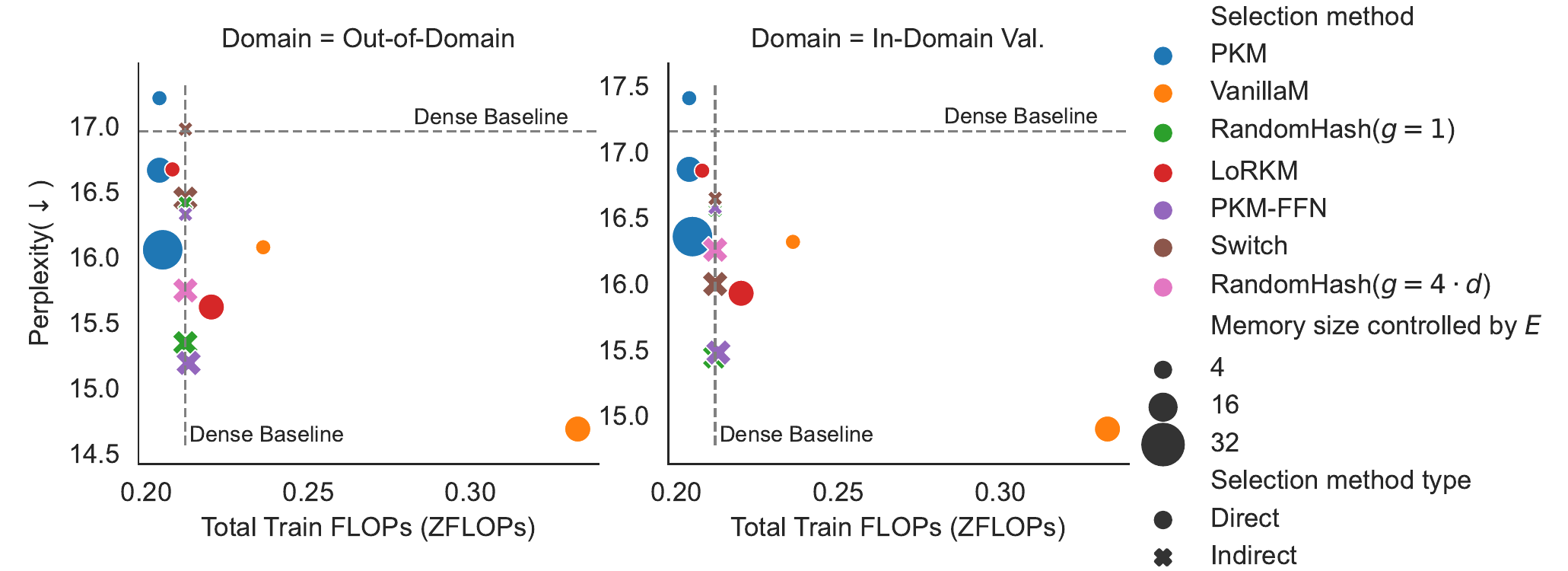}
\caption{FLOPs-Perplexity trade-off different models where direct/indirect methods are further distinguished by model name.
}
\label{fig:355m_pareto:detailed}
\end{figure*}
\pdfoutput=1
\section{\newMethod}
\label{appendix:newMethod}
\subsection{Rationale to use \texttt{\textbf{Avg}} in \newMethod}
\label{appendix:newMethod:rationale}
We heavily base our choice on experiments with aggregators in VanillaM (in Table \ref{tab:oracle-aggregator}). From the experiments with average absolute value (after GeLU), we hypothesized that a positive feature is good at predicting the value of a label/token against all others. In contrast, a negative value is good at negating the prediction of a single token. As such, positive features are more predictive than negative ones. Although the situation might be different for \newMethod (before GeLU), we expect the selection will only be affected more because of the larger impact of negative value. 

Also, we consider the experiment with max-pooled hidden states (i.e., Max($ \cdot $)). This experiment shows that a memory block hardly has a single key-value cell that dominates over others since Max($ \cdot $) underperforms Avg($ \cdot $) and Avg($| \cdot |$). What makes it worse, the max operation will overlook lots of hidden states at selection, but the overlooked hidden states still contribute to the computation. In contrast, the performance increases when we consider the average (or average of the absolute values) where every hidden state contributes to the decision. Although the situation is slightly different in \newMethod , the “max-pooled” version of \newMethod will only overestimate the hidden states information even more, and the aggregated value won’t be indicative of the hidden states used for computation.

The last consideration we have is that the average function is linear. When we select experts, we use the dot product between input and averaged keys. Due to the linearity, this value is equivalent to taking the dot product between the input and every key and taking the average (See Appendix \ref{appendix:newMethod:analysis}). Thus, using this design choice saves a great amount of computation compared with VanillaM, while keeping the neural memory analogy.

\subsection{\newMethod analysis through comparison with \oracle}
\label{appendix:newMethod:analysis}

\newMethod essentially applies an average pooling to the unfactorized $\mathbf{K}^{g}$ to create representation of each block. Due to the linearity of averaging, the operation $\mathbf{e}_i \cdot \mathbf{x}$ is equivalent to calculate the average of dot products within a block before GeLU and select blocks with the average of dot products:
\begin{align*}
    \mathbf{e}_i \cdot \mathbf{x} 
    & = \left(\frac{1}{g}\cdot \sum_{j=0}^{g-1} \mathbf{k}^{(i)}_j \right) \cdot \mathbf{x} \\
    & = \frac{1}{g} \cdot \sum_{j=0}^{g-1} \left(\mathbf{k}^{(i)}_j \cdot \mathbf{x}\right) \\
    & = \texttt{Avg}\left(\mathbf{x} \cdot \left(\mathbf{K}^{(i)}\right)^\top \texttt{,dim=0}\right) \;\;\; (\newMethod)
\end{align*}

In contrast,  \oracle uses average \textit{after} GeLU(\textsection \ref{sec:framework_analysis:granularity}):
$$\frac{1}{g} \sum_{j=0}^{g-1} \text{GeLU}\left(\mathbf{k}^{(i)}_j \cdot \mathbf{x}\right)\;\;\; (\oracle)$$
Because GeLU is a non-linear function, average from \newMethod could be shared across tokens. In contrast, \oracle can't, and thus making \newMethod efficient.

In Fig \ref{fig:new_method:granularity_with_oracle}, we experiment both methods with various $g$. We observe when $g$ decreases from $4096$, the perplexity of \newMethod drops more drastically than \oracle. We believe this observation highlights the impact of GeLU. Because $\displaystyle \lim_{x \to -\infty} \text{GeLU}(x) = 0$, it protects the average in \oracle from some very negative values. Thus, \newMethod with larger $g$ included more and potentially very negative values to average over, and thus leads to worse choices than ones made by \oracle. On the other hand, when $g$ decreases, this ``negative value'' problem is mitigated. When there are more blocks available for selection (smaller $g$), because negative dot products affects \newMethod more, it prefers blocks with more or very positive dot products; whereas, \oracle is protected from negative value so it fails to detect those blocks. Therefore, \newMethod with $g \leq 256$ could achieve an even better perplexity.

\subsection{Why \newMethod with load balancing should work?}
\label{appendix:newMethod:load-balancing-rationale}
\begin{figure*}[th]
    \centering
     \includegraphics[scale=0.7,trim={1em 1.4em 1em 0.5em},clip]{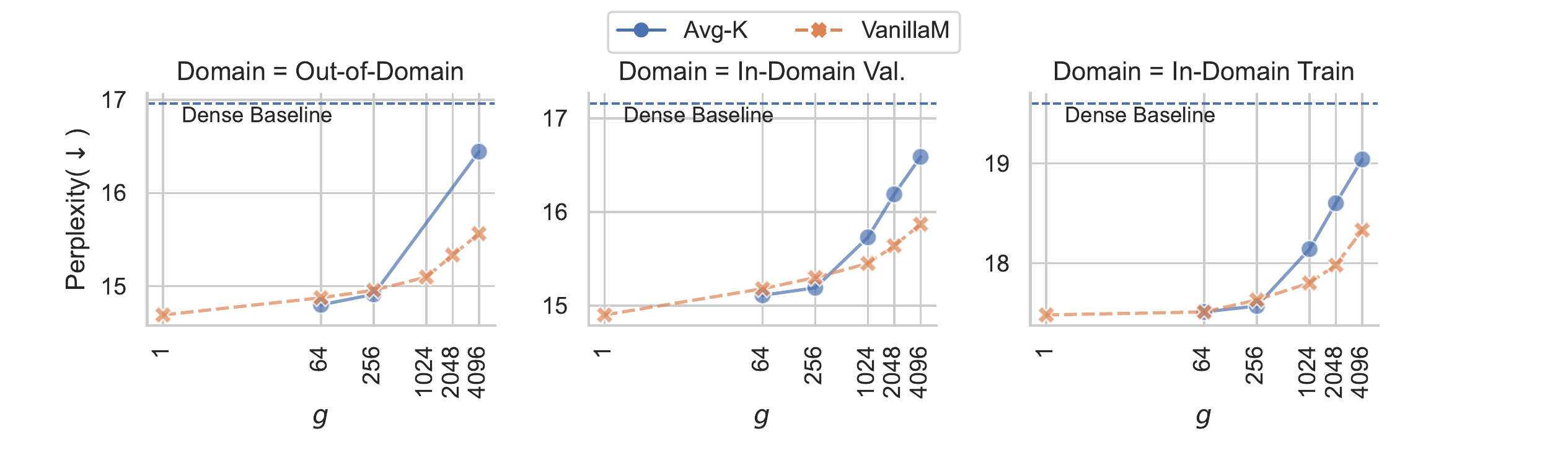}
    \caption{ Perplexity performance (lower the better) of \newMethod and \oracle across various $g$. We observe large drop in perplexity when $g$ decreases in \newMethod and less so in \oracle; and \newMethod slightly outperform \oracle with $g\leq 256$.
    }
    \label{fig:new_method:granularity_with_oracle}
\end{figure*}

Comparing \oracle and \newMethod, one would expect \newMethod to be greatly affected by extremely negative hidden states (before GeLU). Yet, the final model with \newMethod could even outperform VanillaM with the same block size (Fig. \ref{fig:new_method:granularity_with_oracle}). This means the model will accommodate small design changes. 

Additionally, the requirement of a load balancing loss is determined by the sparsity of gradients. If one “expert” gets updated with gradients while other “experts” are starved then the one expert will be selected all of the time leading to a form of mode collapse. In fact, with the same memory block size ($g  = 4096$), we are surprised to observe \newMethod (w/o load balancing loss; row 6 in Table \ref{tab:new_method:comparison}) could still perform on par with Switch (w/ load balancing; row 4 in Table \ref{tab:new_method:comparison}). As our load balancing analysis suggests in Appendix \ref{appendix:newMethod:load-balance}, when the number of experts is small, the mode collapse issue in \newMethod is severe. This makes us more confident that \newMethod will perform better with standard load balancing loss added. 

Therefore, we believe that with the loss added, the model will accommodate and could still perform competitively.

\subsection{\newMethod load balancing analysis}
\label{appendix:newMethod:load-balance}
\begin{figure*}[th]
\centering
    \begin{subfigure}{1\linewidth}
         \centering
          \includegraphics[width=1\linewidth]{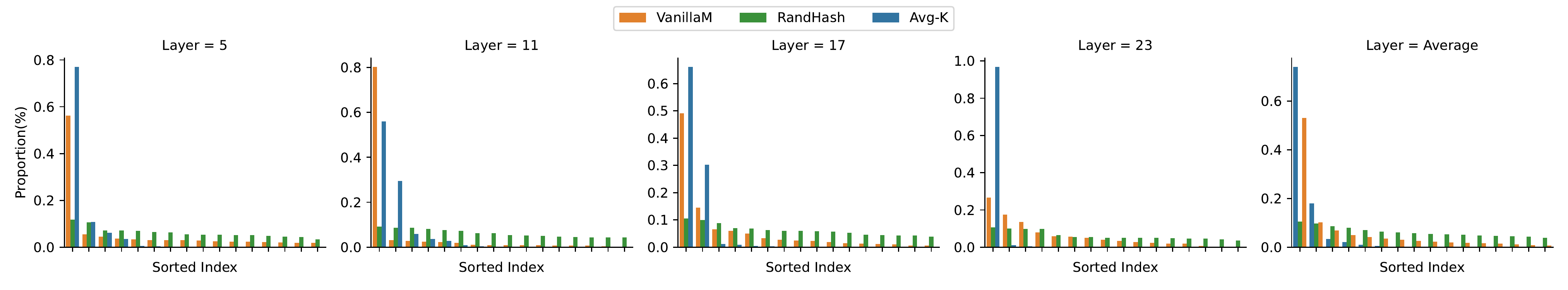}
         \caption{$g=4096$}
         \label{fig:new_method:load-balance:g=4096}
    \end{subfigure}
    \begin{subfigure}{1\linewidth}
         \centering
          \includegraphics[width=1\linewidth]{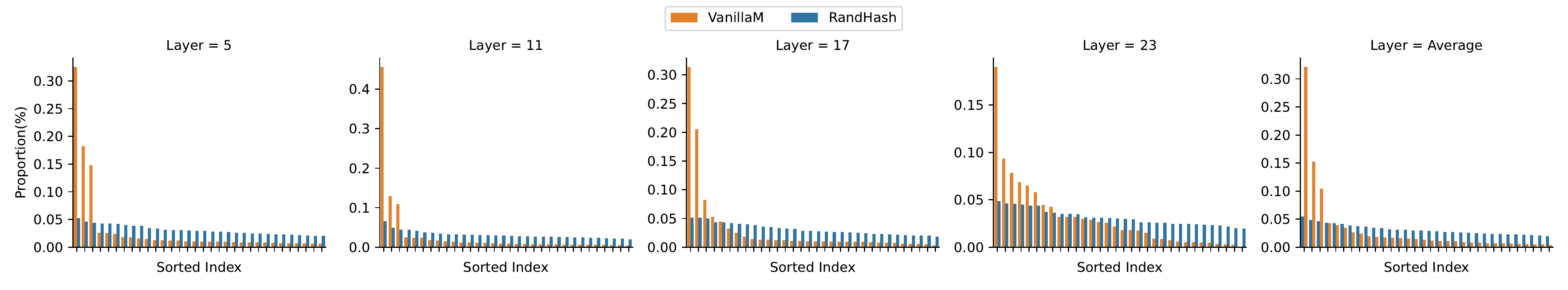}
         \caption{$g=2048$}
         \label{fig:new_method:load-balance:g=2048}
    \end{subfigure}
    \begin{subfigure}{1\linewidth}
         \centering
          \includegraphics[width=1\linewidth]{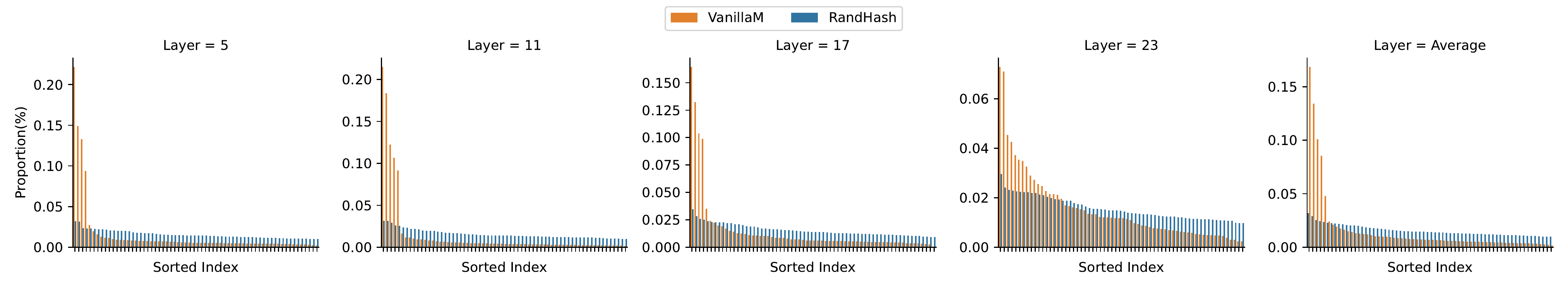}
         \caption{$g=1024$}
         \label{fig:new_method:load-balance:g=1024}
    \end{subfigure}
    \begin{subfigure}{1\linewidth}
         \centering
          \includegraphics[width=1\linewidth]{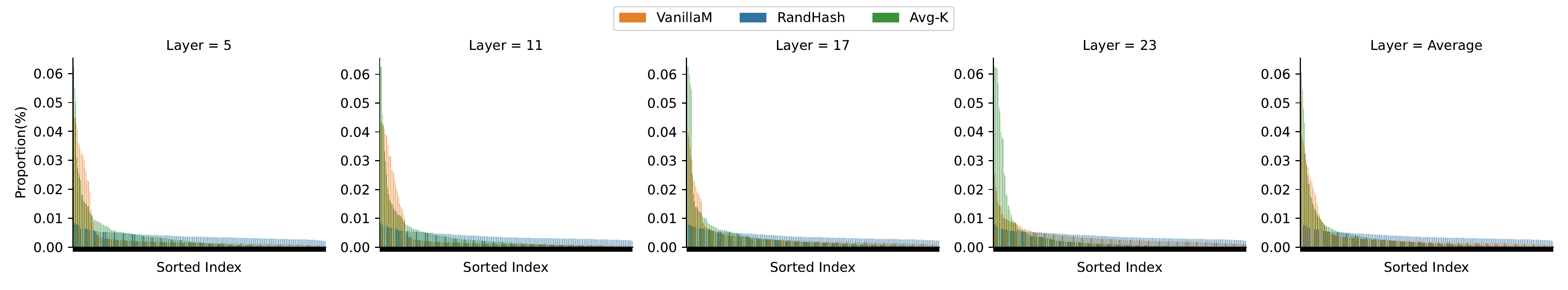}
         \caption{$g=256$}
         \label{fig:new_method:load-balance:g=256}
    \end{subfigure}
    \begin{subfigure}{1\linewidth}
         \centering
          \includegraphics[width=1\linewidth]{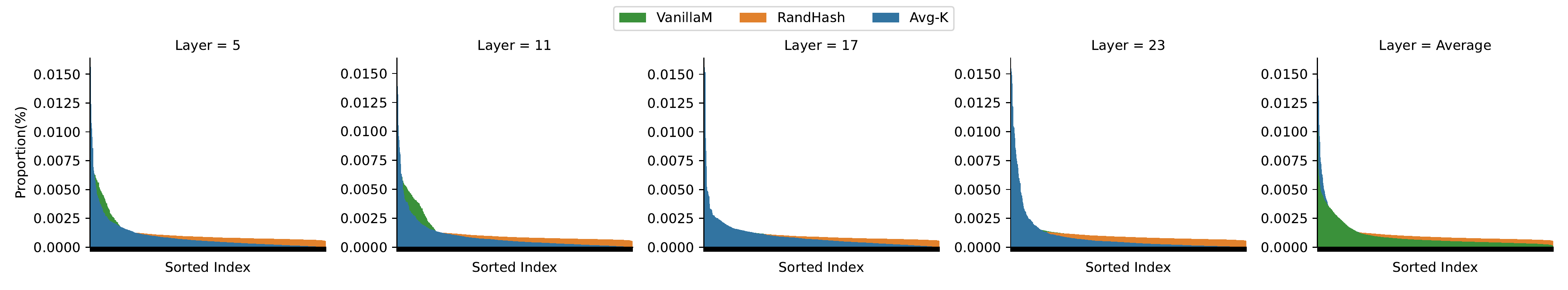}
         \caption{$g=64$}
         \label{fig:new_method:load-balance:g=64}
    \end{subfigure}
\caption{Load balancing of \oracle, \hash, \newMethod. The height of bar represents the proportion of memory block usage with which the memory block are sorted (in descending order).}
\label{fig:new_method:load-balance}
\end{figure*}

On the same validation set as used in \textsection \ref{appendix:granularity}, we also conduct a load balancing analysis of memory blocks. Fig. \ref{fig:new_method:load-balance} shows that \newMethod and \oracle disproportionally used some memory blocks.
\pdfoutput=1
\section{Preliminary study for related work}
\label{appendix:related}

\subsection{Terraformer analysis}
\label{appendix:related:terraformer}



\paragraph{Controller in Terraformer} \citet{sparse_is_enough} uses a controller to score all memory cells and pre-select a subsets --- $\text{Controller}(\mathbf{x})$ --- for computation. 
\begin{align*}
     \mathbf{y} & = \sum_{i\in \text{Controller}(\mathbf{x})} f(\mathbf{x} \cdot \mathbf{k}_i) \cdot \mathbf{v}_i
\end{align*}

This is closest to our \pkmFfn, since their controller is essentially a gate with low-rank key table in \lowRank --- $\mathbf{g}(\mathbf{x}) = (\mathbf{x} \cdot \mathbf{D}) \cdot (\mathbf{K}')^\top$, where $\mathbf{D} \in \mathbb{R}^{d\times d_\ell}$, $\mathbf{K} \in \mathbb{R}^{d_m\times d_\ell}$, and $d_\ell \ll d$. 
The difference is that they additionally assume the estimation from gate (and memory) could be seen as chunked into blocks and only select top-1 memory cell scored by the controller from each blocks:
\begin{align*}
    \mathbf{y} = \sum_{i=0}^{B-1} \mathbf{g}(\mathbf{x})^{(i)}_{j^*} \cdot f(\mathbf{x} \cdot \mathbf{k}_{j^*}^{(i)}) \cdot \mathbf{v}_{j^*}^{(i)} \\
\end{align*}
$\text{where }j^* = \argmax_j \mathbf{g}(\mathbf{x})^{(i)}_j$. 
Therefore, their number of active memory cells $k$ is equal to $d_m / g$.

\begin{figure*}[th]
    \centering
    \begin{subfigure}{1\linewidth}
     \centering
      \includegraphics[width=1\linewidth]{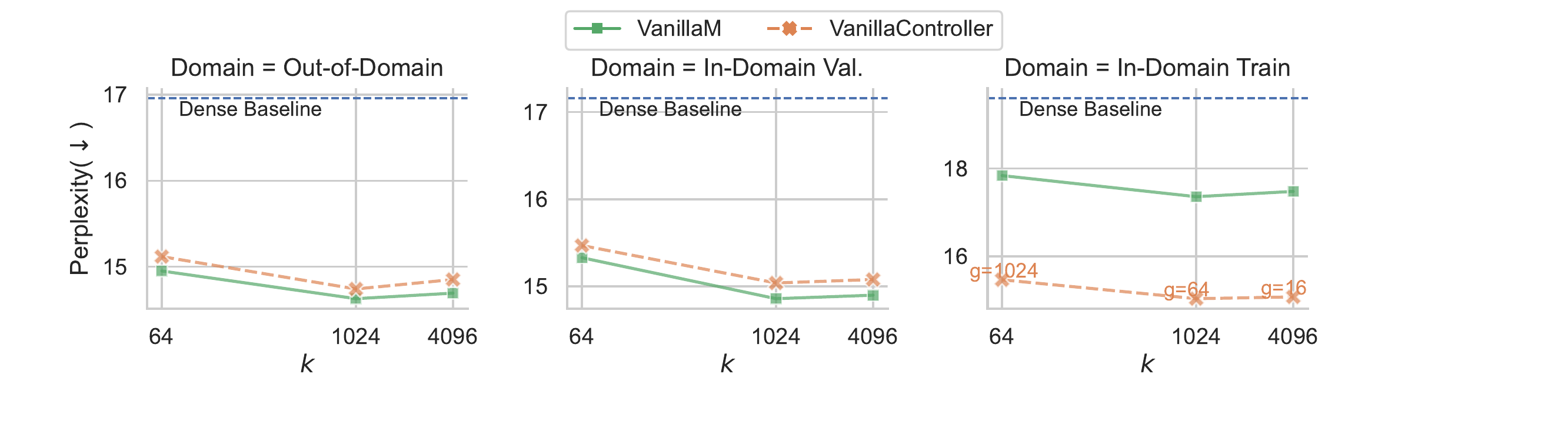}
     \caption{Average perplexity performance (lower the better) }
     \label{fig:terraformer:avg}
    \end{subfigure}
    \\
    \begin{subfigure}{1\linewidth}
     \centering
       \includegraphics[width=1\linewidth]{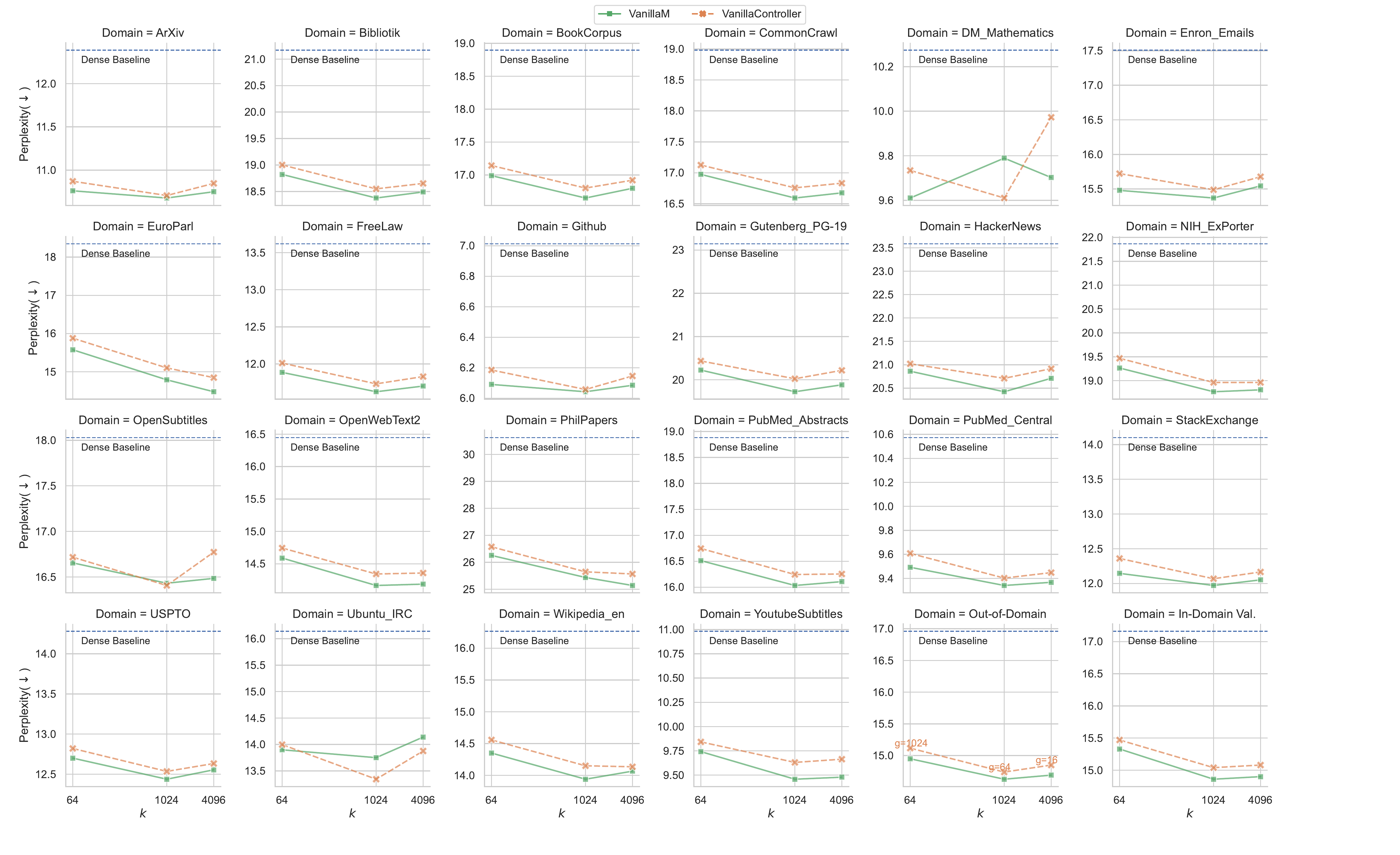}
     \caption{Performance on individual domain in PILE perplexity (lower the better)}
     \label{fig:terraformer:domain}
    \end{subfigure}
\caption{Perplexity performance (lower the better) of \oracle(g=1) and \oracleTerraformer with $E=16$. }
\label{fig:terraformer}
\end{figure*}

Similar to our contrastive pair of \pkmFfn and \oracle, we hypothesize a ``vanilla'' version of their methods. Memory is chunked into blocks of size $g$ --- $\mathbf{K}^g = [\mathbf{K}^{(0)}; \cdots; \mathbf{K}^{(B-1)} ]$ and similarly for $\mathbf{V}^g$. Then, one chooses the top-1 with $\mathbf{x} \cdot (\mathbf{K}^{(i)})^\top$. We call it \textbf{\oracleTerraformer}.

\begin{align*}
    \mathbf{y} = \sum_{i=0}^{B-1} f(\mathbf{x} \cdot \mathbf{k}_{j^*}^{(i)}) \cdot \mathbf{v}_{j^*}^{(i)} 
\end{align*}
$\text{where } j^* = \argmax_j \mathbf{x} \cdot (\mathbf{K}^{(i)})^\top$.
In Fig. \ref{fig:terraformer}, we compare \oracleTerraformer to \oracle with $g=1$, because the \textit{actual} section is at the level of $g=1$. We set $k$ in \oracle to the one determined by equation above. We observe \oracle outperforms \oracleTerraformer. Although the controller design as a gating function is justified (\textsection \ref{sec:framework_analysis:selection}), the decision choice of ``chunking memory but only select the best memory cells'' seems unmotivated. Thus, we exclude this design setup from our analysis.

\subsection{ANN}
\label{appendix:related:ann}
\begin{figure*}[th]
\centering
    \begin{subfigure}{1\linewidth}
         \centering
          \includegraphics[width=1\linewidth]{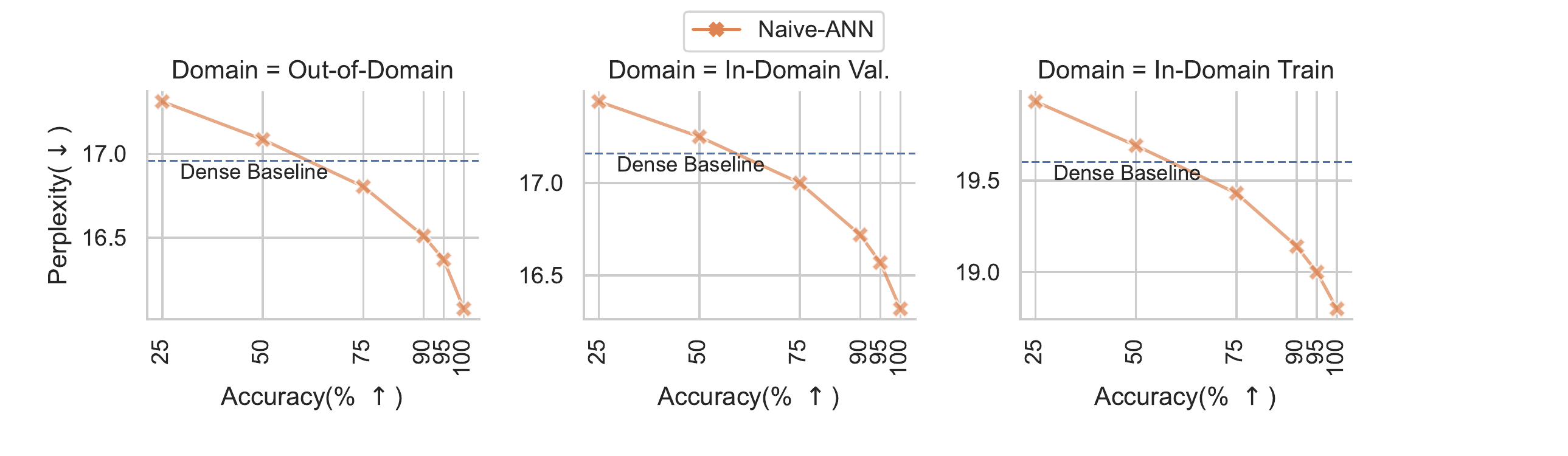}
         \caption{Average perplexity performance (lower the better) }
         \label{fig:naive-ann:avg}
    \end{subfigure}
    \begin{subfigure}{1\linewidth}
         \centering
         \includegraphics[width=1\linewidth]{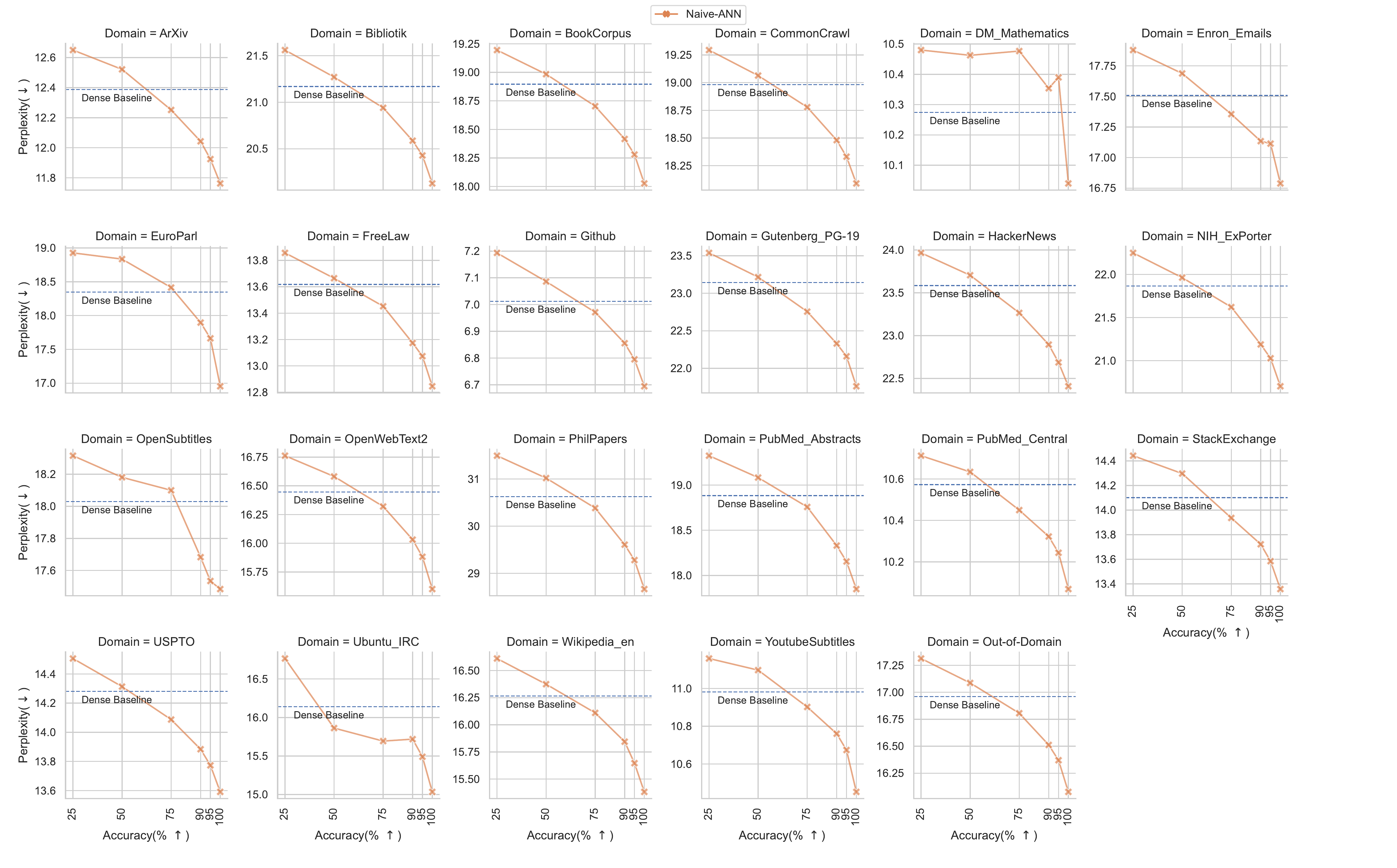}
         \caption{Performance on individual domain perplexity in PILE (lower the better)}
         \label{fig:naive-ann:domain}
    \end{subfigure}
\caption{Perplexity performance (lower the better) of Naive-ANN accuracy with $E=4$. }
\label{fig:naive-ann}
\end{figure*}

Since ANN is an approximation to exact search, we propose to randomly sabotage \oracle, which uses the exact search. Given a $k$, we randomly swap $n$\% of the top-$k$ of memory coefficient $\mathbf{m}$ (exact search results) with non-top-$k$ values (during training and validation), and has accuracy $(100-n)\%$ We call it \textbf{Naive-ANN}. This is meant to set up a random baseline for ANN, because different ANN techniques might make systematic mistakes, rather than a random one. However, we believe this could still serve as a proxy and shed light on how it affects performance. As we see in Fig. \ref{fig:naive-ann}, the model quality is sensitive to the quality of ANN. 

In our preliminary study, we found building data structure after every update is expensive. This leads to some critical drawback when we apply the techniques to model parameter. Although one could amortize the cost by periodically building, the outdated data structure will lead to lower accuracy. If one chooses a hyperparameter that leads to higher quality, the cost of preprocessing and the corresponding search will be even higher. What makes it worse, the current ANN methods' search either don't support speedup by using GPU, or is not very well-integrated with GPUs --- slower than calculating the exact dot product with CUDA kernel.

\pdfoutput=1
\begin{table*}[th]
\centering
\setlength{\tabcolsep}{3pt}
\def\arraystretch{1.2}      

\begin{tabular}{ll|c|cccccccc}
\toprule
   &   \multirow{2}{*}{\begin{tabular}[c]{@{}c@{}}
    Selection method \\
    Type
\end{tabular}}   & &   &  \multicolumn{4}{c}{\multirow{2}{*}{Direct}} &  \multicolumn{3}{c}{\multirow{2}{*}{Indirect}} \\
{} &  &           &       &       &                &                &        &    &          &           \\
\comment{\cline{1-2}} \cline{5-8} \cline{10-11}
   &   \multirow{2}{*}{Selection method}     &  \multirow{2}{*}{\begin{tabular}[c]{@{}c@{}}
    Dense \\ Baseline
\end{tabular}} &  &  \multicolumn{3}{c}{\multirow{2}{*}{PKM}}  & \multirow{2}{*}{VanillaM} & &  \multirow{2}{*}{PKM-FFN} &  \multirow{2}{*}{RandHash} \\
{} &  &             &     &       &                &                &           &          &           \\
\comment{\cline{1-2}} \cline{5-7}
{} & $E$ &     1     &    &   16   &    32    &   32    &    16  &  &     16   &     16    \\
{} & $k$ &     4096   &    &   4096    &   4096    &     8192     &     4096  &   &   4096    &    4096   \\
\midrule
\multirow{22}{*}{\parbox[t]{2mm}{\multirow{3}{*}{\rotatebox[origin=c]{90}{Out-of-Domain}}}}
          & ArXiv &           12.39 & & 12.20 &          11.82 &          11.89 &     \textbf{10.75} &  &   \textbf{11.05} &     11.22 \\
          & Bibliotik &           21.17 & & 20.82 &          20.15 &          20.25 &     \textbf{18.49} &  &   \textbf{19.11} &     19.33 \\
          & BookCorpus &           18.90 & & 18.61 &          18.09 &          18.19 &     \textbf{16.80} &  &   \textbf{17.26} &     17.49 \\
          & CommonCrawl &           18.98 & & 18.68 &          18.09 &          18.20 &     \textbf{16.68} &  &   \textbf{17.23} &     17.40 \\
          & DM\_Mathematics &           10.27 & & 10.34 &          10.05 &          10.28 &     \textbf{ 9.70} &  &    \textbf{9.72} &      9.91 \\
          & Enron\_Emails &           17.51 & & 17.23 &          16.67 &          16.68 &     \textbf{15.55} & &    \textbf{15.90} &     16.18 \\
          & EuroParl &           18.35 & & 17.79 &          17.01 &          17.05 &     \textbf{14.48} &  &   15.50 &     \textbf{15.03} \\
          & FreeLaw &           13.62 & & 13.34 &          12.84 &          12.93 &     \textbf{11.70} &  &   \textbf{12.11} &     12.29 \\
          & Github &            7.01 & &  6.91 &           6.67 &           6.68 &      \textbf{6.08} &   &   \textbf{6.25} &      6.37 \\
          & Gutenberg\_PG-19 &           23.14 & & 22.61 &          21.83 &          22.03 &     \textbf{19.88} &  &   \textbf{20.74} &     21.07 \\
          & HackerNews &           23.58 & & 23.35 &          22.44 &          22.62 &     \textbf{20.71} &  &   \textbf{21.36} &     21.76 \\
          & NIH\_ExPorter &           21.87 & & 21.45 &          20.59 &          20.77 &     \textbf{18.81} &  &   \textbf{19.48} &     19.69 \\
          & OpenSubtitles &           18.03 & & 17.99 &          17.46 &          17.44 &     \textbf{16.48} &  &   \textbf{16.84} &     17.10 \\
          & OpenWebText2 &           16.45 & & 16.15 &          15.60 &          15.68 &    \textbf{ 14.19} &  &   \textbf{14.73} &     14.74 \\
          & PhilPapers &           30.63 & & 30.02 &          28.50 &          28.74 &     \textbf{25.14} &  &   \textbf{26.44} &     26.60 \\
          & PubMed\_Abstracts &           18.88 & & 18.50 &          17.71 &          17.92 &     \textbf{16.11} &   &  \textbf{16.75} &     16.90 \\
          & PubMed\_Central &           10.57 & & 10.40 &          10.08 &          10.14 &      \textbf{9.37} &   &   \textbf{9.66} &      9.71 \\
          & StackExchange &           14.10 & & 13.85 &          13.37 &          13.45 &     \textbf{12.05} &   &  \textbf{12.46} &     12.72 \\
          & USPTO &           14.28 & & 14.07 &          13.58 &          13.68 &     \textbf{12.55} &  &   \textbf{12.96} &     13.09 \\
          & Ubuntu\_IRC &           16.14 & & 15.40 &          14.95 &          15.08 &     \textbf{14.14} &  &   \textbf{14.35} &     14.62 \\
          & Wikipedia\_en &           16.26 & & 16.03 &          15.33 &          15.48 &     \textbf{14.07} &  &   \textbf{14.51} &     14.59 \\
          & YoutubeSubtitles &           10.98 & & 10.86 &          10.41 &          10.41 &      \textbf{9.48} &  &    9.87 &      \textbf{9.77} \\
\midrule
          & Average &           16.96 & &  16.66 &          16.06 &          16.16 &     \textbf{14.69} &  &   \textbf{15.19} &     15.35 \\
\bottomrule
\end{tabular}
\caption{Detailed out-of-domain perplexity for Table \ref{tab:355m_full_key}. Best two performance on each domain is in bold. Relative ranking on each domain generally follows the relative ranking by averaged performance (i.e. last row).}
\label{tab:355m_full_key:OOD}
\end{table*}
\pdfoutput=1
\begin{table*}[th]
\centering
\setlength{\tabcolsep}{4pt}
\def\arraystretch{1.2}      

\begin{tabular}{ll|cccccccc}
\toprule
      & \multirow{2}{*}{Selection method} & \multirow{2}{*}{\begin{tabular}[c]{@{}c@{}}
    Dense \\ Baseline
\end{tabular}} & \multicolumn{2}{c}{\multirow{2}{*}{RandHash}} & \multirow{2}{*}{Switch} & \multirow{2}{*}{PKM-FFN} & \multicolumn{3}{c}{\multirow{2}{*}{\newMethod}} \\
      &   &                      &      &      &    &        &   &    &     \\ \cline{4-5} \cline{8-10}
      & $g$ &                  1    &     4096 &  1    &   4096 &    1    &  4096 &  256  &  64   \\
\midrule
  \multirow{22}{*}{\parbox[t]{2mm}{\multirow{3}{*}{\rotatebox[origin=c]{90}{Out-of-Domain}}}}    
      & ArXiv &                 12.39 &    11.42 & 11.22 &  12.32 &   11.05 & 12.09 & 10.99 & \textbf{10.85} \\
      & Bibliotik &                 21.17 &    19.84 & 19.33 &  19.87 &   19.11 & 20.49 & 18.70 & \textbf{18.61} \\
      & BookCorpus &                 18.90 &    17.94 & 17.49 &  17.85 &   17.26 & 18.33 & 16.86 & \textbf{16.82} \\
      & CommonCrawl &                 18.98 &    17.84 & 17.40 &  17.70 &   17.23 & 18.42 & 16.89 & \textbf{16.81} \\
      & DM\_Mathematics &                 10.27 &    10.22 &  9.91 &  10.63 &    9.72 & 10.25 &  9.62 &  \textbf{9.61} \\
      & Enron\_Emails &                 17.51 &    16.70 & 16.18 &  17.36 &   15.90 & 17.20 & 15.65 & \textbf{15.55} \\
      & EuroParl &                 18.35 &    15.55 & 15.03 &  19.63 &   15.50 & 17.38 & 15.32 & \textbf{15.13} \\
      & FreeLaw &                 13.62 &    12.56 & 12.29 &  12.80 &   12.11 & 13.20 & 11.84 & \textbf{11.77} \\
      & Github &                  7.01 &     6.51 &  6.37 &   6.96 &    6.25 &  6.89 &  6.18 &  \textbf{6.12} \\
      & Gutenberg\_PG-19 &                 23.14 &    21.55 & 21.07 &  21.81 &   20.74 & 22.36 & 20.07 & \textbf{19.98} \\
      & HackerNews &                 23.58 &    22.30 & 21.76 &  22.59 &   21.36 & 22.95 & 20.82 & \textbf{20.65} \\
      & NIH\_ExPorter &                 21.87 &    20.22 & 19.69 &  20.99 &   19.48 & 21.09 & 19.09 & \textbf{19.01} \\
      & OpenSubtitles &                 18.03 &    17.33 & 17.10 &  17.31 &   16.84 & 17.74 & \textbf{16.62} & 16.66 \\
      & OpenWebText2 &                 16.45 &    15.16 & 14.74 &  15.51 &   14.73 & 15.87 & 14.48 & \textbf{14.39} \\
      & PhilPapers &                 30.63 &    27.53 & 26.60 &  30.84 &   26.44 & 29.51 & 25.90 & \textbf{25.72} \\
      & PubMed\_Abstracts &                 18.88 &    17.39 & 16.90 &  18.36 &   16.75 & 18.24 & 16.34 & \textbf{16.33} \\
      & PubMed\_Central &                 10.57 &     9.94 &  9.71 &  10.28 &    9.66 & 10.30 &  9.50 &  \textbf{9.45} \\
      & StackExchange &                 14.10 &    13.04 & 12.72 &  13.78 &   12.46 & 13.73 & 12.23 & \textbf{12.13} \\
      & USPTO &                 14.28 &    13.41 & 13.09 &  13.54 &   12.96 & 13.89 & 12.73 & \textbf{12.62} \\
      & Ubuntu\_IRC &                 16.14 &    14.95 & 14.62 &  14.78 &   14.35 & 15.50 & 14.26 & \textbf{13.59} \\
      & Wikipedia\_en &                 16.26 &    14.98 & 14.59 &  15.67 &   14.51 & 15.73 & 14.23 & \textbf{14.12} \\
      & YoutubeSubtitles &                 10.98 &    10.06 &  9.77 &  11.25 &    9.87 & 10.59 &  9.73 &  \textbf{9.67} \\
\midrule
      & Average &                 16.96 &    15.75 & 15.35 &  16.45 &   15.19 & 16.44 & 14.91 & \textbf{14.80} \\
\bottomrule
\end{tabular}
\caption{Detailed out-of-domain perplexity for Table \ref{tab:new_method:comparison}. Best performance on each domain is in bold. Relative ranking on each domain generally follows the relative ranking by averaged performance (i.e. last row).}
\label{tab:new_method:comparison:OOD}
\end{table*}

\end{document}